\documentclass{article}

\usepackage{microtype}
\usepackage{graphicx}
\usepackage{subcaption}
\usepackage{booktabs} 

\usepackage{hyperref}

\usepackage{algpseudocode}

\usepackage{wrapfig}
\usepackage[preprint]{icml2026}

\usepackage{amsmath}
\usepackage{amssymb}
\usepackage{mathtools}
\usepackage{amsthm}

\usepackage{url}
\usepackage{amsmath}      
\usepackage{amsthm}       
\usepackage{multirow}
\usepackage[table]{xcolor} 
\usepackage{tabularx,array} 
\usepackage{tikz}

\usepackage[capitalize,noabbrev]{cleveref}

\theoremstyle{plain}
\newtheorem{theorem}{Theorem}[section]

\theoremstyle{definition}

\theoremstyle{remark}

\usepackage[textsize=tiny]{todonotes}

\icmltitlerunning{Dataset Distillation via Committee Voting}
\newcommand{\cjc}[1]{{\color{black}#1}}
\newcommand{\name}{\textbf{CV-DD}}
\DeclareMathOperator*{\argmin}{arg\,min}

\begin{document}

\twocolumn[
  \icmltitle{Dataset Distillation via Committee Voting}



  \icmlsetsymbol{equal}{*}

  \begin{icmlauthorlist}
    \icmlauthor{Jiacheng Cui}{yyy}
    \icmlauthor{Zhaoyi Li}{yyy}
    \icmlauthor{Xiaochen Ma}{comp}
    \icmlauthor{Xinyue Bi}{yyy}
    \icmlauthor{Yaxin Luo}{yyy}
    \icmlauthor{Zhiqiang Shen}{yyy}
  \end{icmlauthorlist}

  \icmlaffiliation{yyy}{Department of Machine Learning, Mohamed bin Zayed University of Artificial Intelligence, Abu Dhabi, UAE}
  \icmlaffiliation{comp}{Independent researcher}

  \icmlcorrespondingauthor{Zhiqiang Shen}{zhiqiang.shen@mbzuai.ac.ae}
  
  \icmlkeywords{Machine Learning, ICML}

  \vskip 0.3in
]



\printAffiliationsAndNotice{}  

\begin{abstract}
Dataset distillation aims to synthesize a compact yet representative dataset that preserves the essential characteristics of the original data for efficient model training. Existing methods mainly focus on improving data-synthetic alignment or scaling distillation to large datasets. In this work, we propose \textbf{C}ommittee \textbf{V}oting for \textbf{D}ataset \textbf{D}istillation ({\bf CV-DD}), an orthogonal approach that leverages the collective knowledge of multiple models to produce higher-quality distilled data. We first establish a strong baseline that achieves state-of-the-art performance through modern architectural and optimization choices. By integrating distributions and predictions from multiple models and generating high-quality soft labels, our method captures a broader range of data characteristics, reduces model-specific bias and the impact of distribution shifts, and significantly improves generalization. This voting-based strategy enhances diversity and robustness, alleviates overfitting, and improves post-evaluation performance. Extensive experiments across multiple datasets and IPC settings demonstrate that CV-DD consistently outperforms single- and multi-model distillation methods and generalizes well to non-training-based frameworks and challenging synthetic-to-real transfer tasks. Code is available at \url{https://github.com/Jiacheng8/CV-DD}.
\end{abstract}

\section{Introduction}
\label{sec:intro}

\begin{figure}[t]
    \centering
    \includegraphics[width=0.8\linewidth]{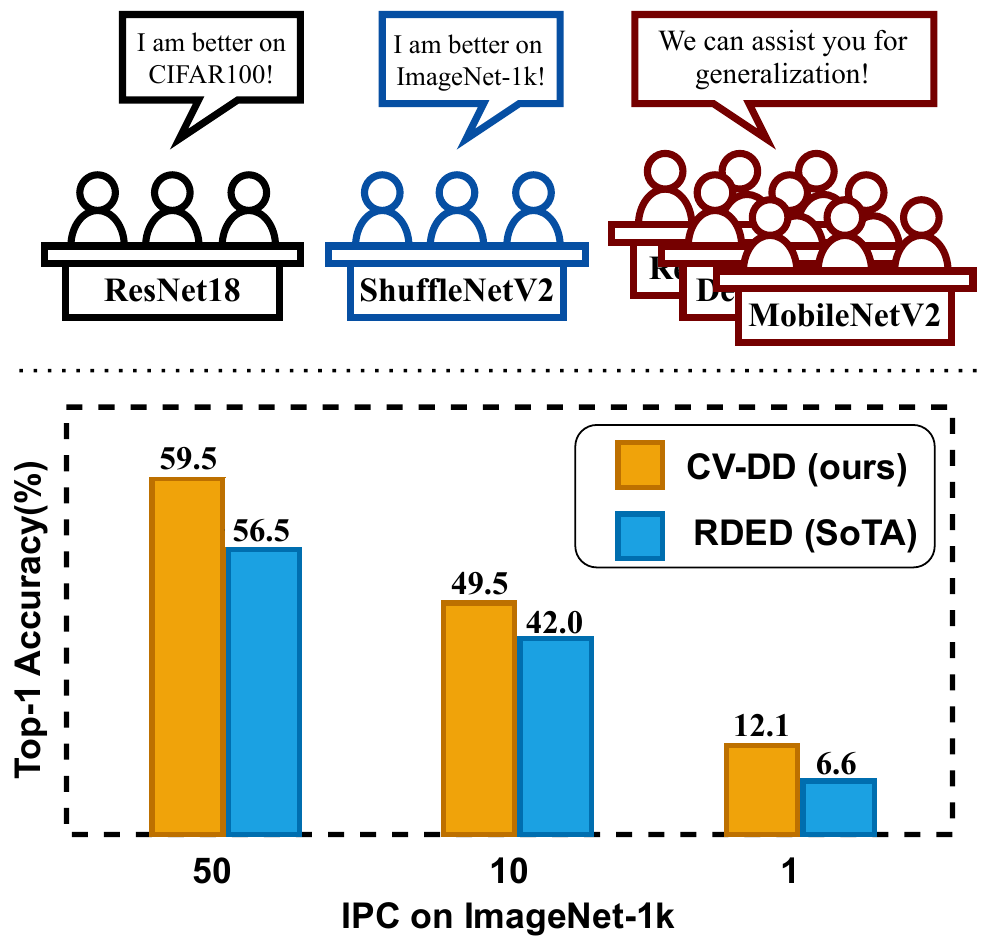} 
    \caption{Top illustrates the motivation of our committee voting-based dataset distillation, highlighting its ability to reduce bias from individual model knowledge. Bottom shows the performance improvement over previous state-of-the-art method RDED~\citep{RDED_2024} on ResNet-18.}
    \vspace{-.23in}
\end{figure}

The rapid growth of large datasets has significantly advanced computer vision and deep learning applications, enabling models to achieve high accuracy and generalization across diverse domains. However, training on massive datasets presents challenges such as high computational cost, memory usage, and long training times, especially for resource-constrained environments. To address these issues, {\em dataset distillation} has emerged as an effective technique to condense large datasets into smaller, representative sets, allowing for efficient model training with minimal performance loss. Despite its promise, a key challenge remains: {\em capturing the essential features of the original data while avoiding overfitting to specific patterns or noise.}

\begin{figure*}[!t]
    \centering
    \includegraphics[width=0.9\textwidth]{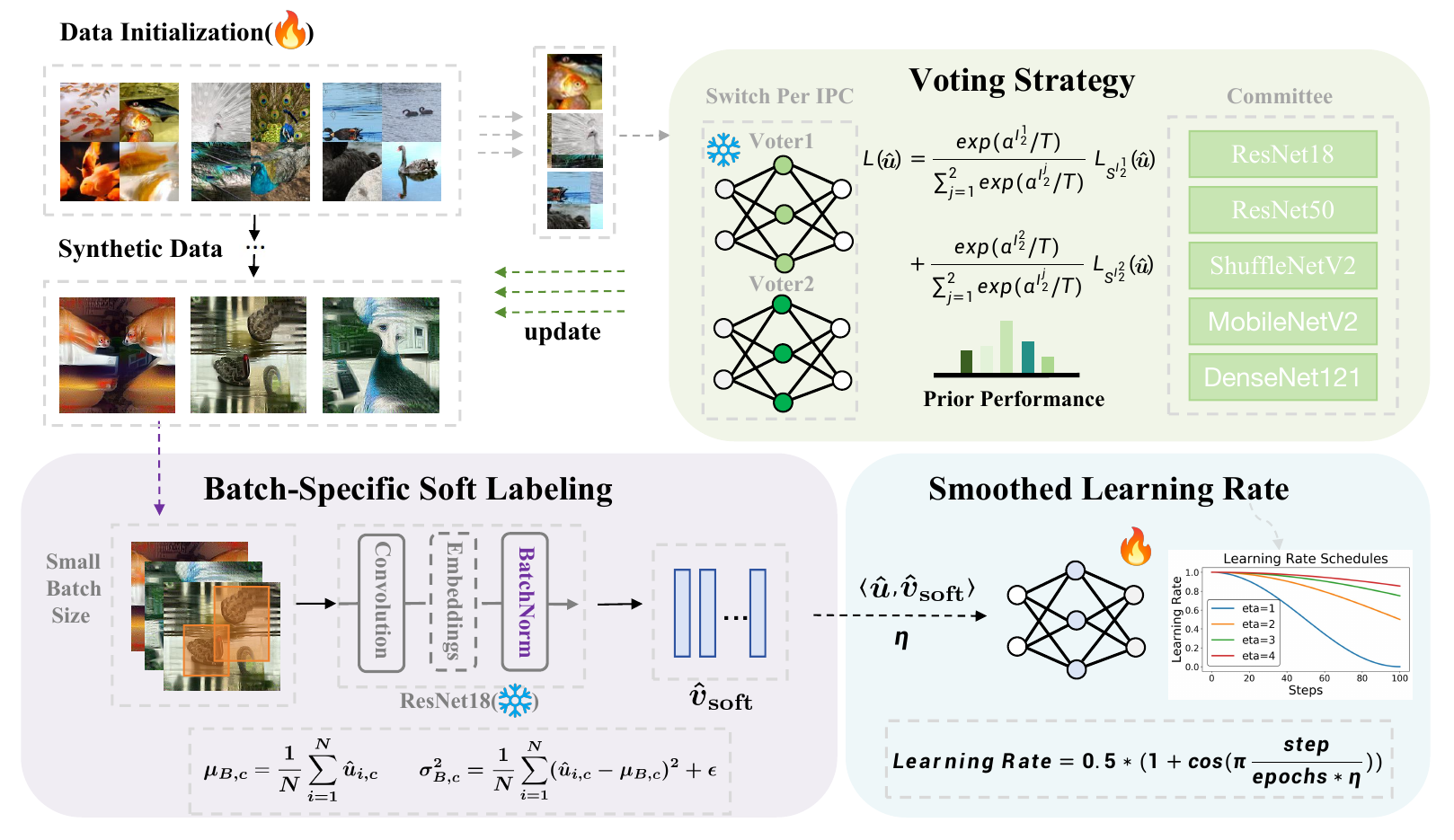}
    \caption{
    The process begins with {\bf Data Initialization} to generate synthetic data from the original data distribution. In {\bf Voting Strategy} section, a committee of models decides on the distributions for synthetic data, where the voting mechanism considers prior performance and calculates a weighted gradient update based on each model's distribution and prediction. {\bf Batch-Specific Soft Labeling} generates soft labels by embedding batch norm statistics from synthetic data batch to mitigate the impact of distribution shift. Finally, a {\bf Smoothed Learning Rate} strategy is applied to the post-training process, adjusting dynamically with a cosine schedule to stabilize training. 
    }
    \label{fig:overview}
    \vspace{-.1in}
\end{figure*}

Prior dataset distillation methods~\citep{sre2l_2024,RDED_2024} rely on single-model frameworks, which may suffer from limited generalization across diverse datasets and architectures and introduce model-specific biases. To address these limitations, we propose {\bf C}ommittee {\bf V}oting for {\bf D}ataset {\bf D}istillation (\name{}), a framework that leverages multiple models’ perspectives to construct a high-quality and balanced distilled dataset. Voting mechanisms have been widely adopted in NLP and LLM tasks~\citep{yang2024llm,qorib-ng-2023-system,wang-plank-2023-actor}. Extending these successes to dataset distillation in computer vision for the first time, \name{} employs a committee-based voting strategy to enhance dataset diversity and quality. Moreover, we identify key pitfalls in recent distillation methods and establish a strong baseline that already achieves state-of-the-art performance. By integrating models with diverse architectures, \name{} captures more comprehensive features and improves the robustness of distilled datasets.

Specifically, \name{} introduces a Prior Performance Guided Voting strategy that aggregates predictions from multiple models to select representative data points, reducing model-specific bias, promoting diversity, and mitigating overfitting. Moreover, \name{} provides fine-grained control through dynamic voting, allowing model weights and thresholds to be adjusted to emphasize specific features or dataset attributes. We further propose Batch-Specific Soft Labeling (BSSL) to mitigate distribution shift between real and synthetic data, improving post-evaluation performance.

Through extensive experiments on CIFAR, Tiny-ImageNet, ImageNet-1K, and its subsets, we show that \name{} consistently outperforms existing distillation methods in accuracy and cross-model generalization. Datasets distilled via committee voting yield superior post-evaluation performance, even under low-data or limited-compute settings. By leveraging complementary strengths of multiple models, \name{} provides a robust and efficient distillation solution. Moreover, CV-DD integrates seamlessly into non-training-based methods such as RDED and generalizes well to challenging scenarios, including synthetic-to-real transfer.

We make the following contributions in this paper: 

\begin{itemize}
\item We propose a novel framework, {\em Committee Voting for Dataset Distillation} (\name{}), which integrates multiple model perspectives to synthesize a distilled dataset that captures informative features and produces high-quality soft labels via batch-specific normalization.
\item By integrating recent advances and refining the framework design and optimization, we establish a strong \name{} baseline that achieves state-of-the-art performance in dataset distillation.
\item Through experiments across multiple datasets, we demonstrate that \name{} improves generalization, mitigates overfitting, and outperforms prior methods in data-limited scenarios, highlighting its effectiveness as a scalable and reliable solution for dataset distillation.
\end{itemize}
\section{Related Work}
\label{sec:related}

\textbf{Dataset Distillation.} Dataset distillation aims to generate a compact, synthetic dataset that retains essential information from a large dataset. This approach facilitates easier data processing, reduces training time, and achieves performance comparable to training with the full dataset. Existing solutions typically fall into five main categories: 1) {\em Meta-Model Matching}: This method optimizes for model transferability on distilled data, involving an outer loop for updating synthetic data and an inner loop for training the network. Examples include \cjc{DD~\citep{wang2018dataset}}, KIP~\citep{nguyen2021dataset}, RFAD~\citep{loo2022efficient}, FRePo~\citep{zhou2022dataset}, LinBa~\citep{deng2022remember}, and MDC~\citep{he2024multisize}. 2)	{\em Gradient Matching}: This approach performs one-step distance matching between models, focusing on aligning gradients. Methods in this category include DC~\citep{zhao2020dataset}, DSA~\citep{zhao2021dataset}, DCC~\citep{lee2022dataset}, IDC~\citep{pmlr-v162-kim22c}, and MP~\citep{zhou2024improve}.	3) {\em Distribution Matching}: Here, the distribution of original and synthetic data is directly matched through a single-level optimization. Approaches include DM~\citep{DBLP:conf/wacv/ZhaoB23}, CAFE~\citep{Wang_2022_CVPR}, HaBa~\citep{liu2022dataset}, KFS~\citep{lee2022dataset}, DataDAM~\citep{Sajedi_2023_ICCV}, FreD~\cite{shin2024frequency}, and GUARD~\citep{xue2024towards}. 4) {\em Trajectory Matching}: This method matches the weight trajectories of models trained on original and synthetic data over multiple steps. Examples include MTT~\citep{MTT}, TESLA~\citep{cui2023scaling}, APM~\citep{chen2023dataset}, and DATM~\citep{guo2024lossless}.
5) {\em Decoupled Optimization with BatchNorm Matching}: SRe$^2$L~\citep{sre2l_2024} first proposes to decouple the model training and data synthesis for dataset distillation. After that, many decoupled methods have been proposed, such as CDA~\citep{yin2023dataset}, LPLD~\citep{LPLD}, DELT~\citep{shen2025delt}, and FADRM~\citep{cui2025fadrm}.

\begin{figure}[t]
    \centering
    \includegraphics[width=1\linewidth]{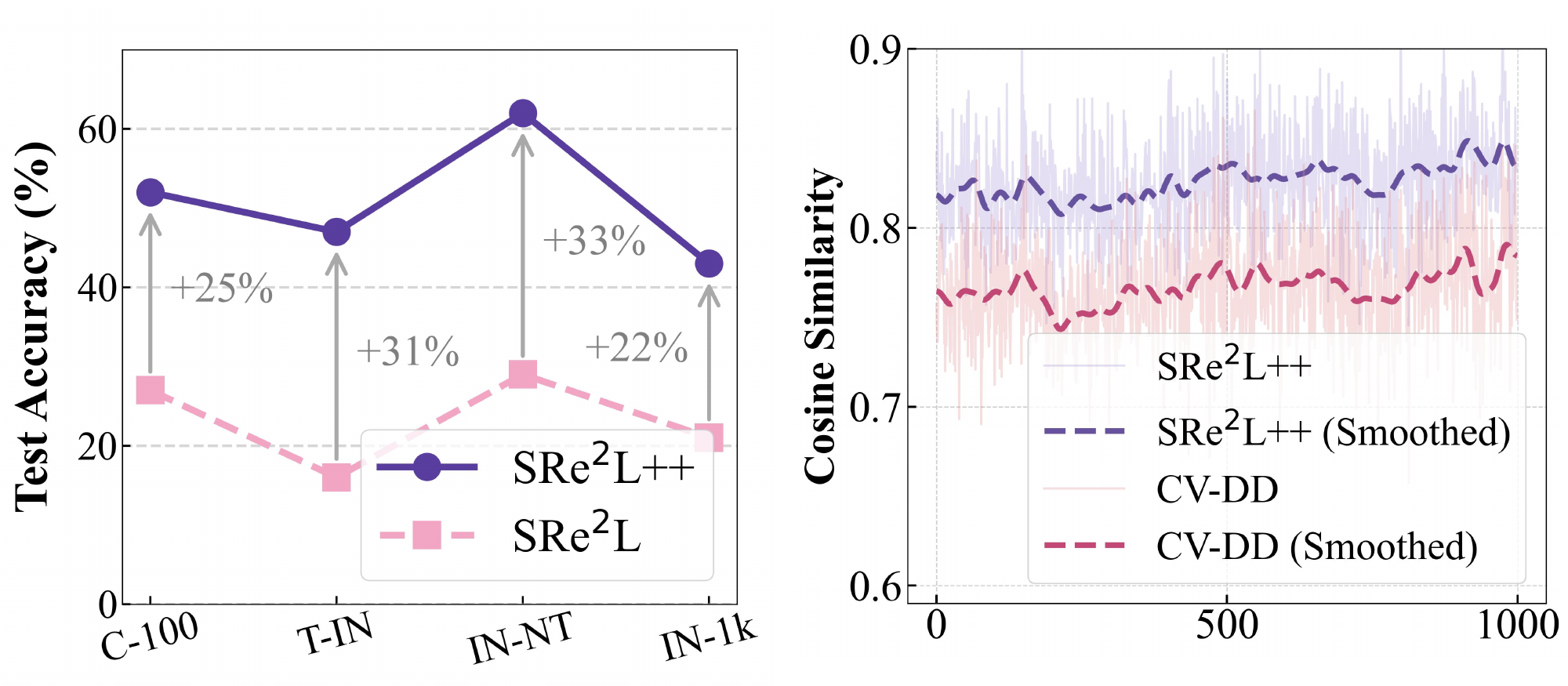}
    \vspace{-.2in}
    \caption{\textbf{Left}: Performance gain of strong baseline.  
    \textbf{Right}: Intra-class cosine similarity on ImageNet-1K (lower is better).}
   \label{fig:performance_and_cos_sim}
   \vspace{-.15in}
\end{figure}

\noindent\textbf{Ensemble Multi-Model Dataset Distillation.}
Ensemble-based methods in dataset distillation aim to leverage multiple models to enhance distilled data quality and generalization. Thus far, few works have explored this direction, notably the MTT series~\citep{MTT,cui2023scaling,FTD} and G-VBSM~\citep{GBVSM_2024}. MTT employs multiple independently trained teachers and leverages their saved checkpoints to guide distillation via expert trajectories. G-VBSM utilizes matching signals from multiple backbones; however, both methods rely on static ensemble configurations without adaptive weighting. In contrast, our proposed {\em Committee Voting} dynamically adjusts model contributions based on prior performance, yielding more effective distilled datasets and improved results.

\vspace{-0.1in}
\section{Approach}
\label{Dataset_Distillation}

\textbf{Preliminaries.}
The goal of dataset distillation is to create a compact synthetic dataset that retains essential information from the original dataset. Given a labeled dataset \( \mathcal{D} = \{(u_1, v_1), \dots, (u_{|\mathcal{D}|}, v_{|\mathcal{D}|})\} \), we aim to learn a synthetic dataset \( \mathcal{D}_{\text{syn}} = \{(\hat{u}_1, \hat{v}_1), \dots, (\hat{u}_{|\mathcal{D}_{\text{syn}}|}, \hat{v}_{|\mathcal{D}_{\text{syn}}|})\} \), where \( |\mathcal{D}_{\text{syn}}| \ll |\mathcal{D}| \). The objective is to minimize the performance gap between models trained on \( \mathcal{D}_{\text{syn}} \) and those trained on \( \mathcal{D} \):
\begin{equation}
\sup_{(u, v) \sim \mathcal{D}} \left| \mathcal{L} \left( \Phi_{\xi_{\mathcal{D}}}(u), v \right) - \mathcal{L} \left( \Phi_{\xi_{\mathcal{D}_{\text{syn}}}}(u), v \right) \right| \leq \delta,
\end{equation}
where \( \delta \) is the allowable gap. This leads to the following optimization problem:
\begin{equation}
\argmin_{\mathcal{D}_{\text{syn}}, \lvert\mathcal{D}_{\text{syn}}\rvert} \sup_{(u, v) \sim \mathcal{D}} \left| \mathcal{L} \left( \Phi_{\xi_{\mathcal{D}}}(u), v \right) - \mathcal{L} \left( \Phi_{\xi_{\mathcal{D}_{\text{syn}}}}(u), v \right) \right|
\end{equation}
The goal is to synthesize \( \mathcal{D}_{\text{syn}} \) while determining the optimal number of samples per class.

\subsection{Pitfalls of Latest Methods}

\noindent\textbf{Diversity and bias issues.} SRe$^2$L~\citep{sre2l_2024} is an optimization-based method that generates distilled data by aligning Batch Normalization (BN) statistics of synthetic data with those observed during training, while enforcing consistency with ground-truth labels. Its main limitation is the reliance on a single backbone for data generation, which restricts diversity and introduces model-specific bias.

\noindent\textbf{Informativeness issues.} Existing ensemble based dataset distillation methods, such as G-VBSM~\citep{GBVSM_2024}, operate under the assumption that all pre-trained models contribute equally during the distillation process. This uniform weighting scheme fails to account for the varying informativeness of individual models, resulting in a distilled dataset that lacks sufficient representational richness.

\noindent \textbf{Suboptimal soft labels.} Prior generative dataset distillation methods~\citep{GBVSM_2024,sre2l_2024,MTT} overlook the distributional shift between synthetic and original images, leading to suboptimal soft labels that hinder model generalization.

\subsection{Building a Strong Baseline}
\label{sec:strong_baseline}

We first introduce SRe$^2$L++, an improved baseline that achieves \emph{SOTA performance.} Its superiority over SRe$^2$L is illustrated in Fig.~\ref{fig:performance_and_cos_sim}.

\noindent\textbf{Real Image Initialization.}
SRe$^2$L++ replaces Gaussian noise with real images for initialization, which showed improved quality at the same cost~\citep{shao2024elucidating}.

\textbf{Data Augmentation.} SRe$^2$L++ improves performance on low-resolution datasets by incorporating data augmentation (RandomResizedCrop) during synthesis (see Fig.~\ref{fig:augmentation}).

\textbf{Batch-Specific Soft Labeling}: To further enhance the performance, we apply the proposed Batch-Specific Soft Labeling technique, detailed in Section~\ref{bssl}.

\textbf{Smoothed Learning Rate and Smaller Batch Size.} Prior studies~\citep{shao2024elucidating} suggest reducing batch size to increase iterations per epoch, thereby mitigating under-convergence, and using a smoothed learning rate schedule to avoid suboptimal minima. These settings are applied to training-based methods when compatible. In contrast, non-training-based methods such as RDED require larger batches, as unoptimized image crops introduce high variance, making BatchNorm unstable with small batches.

\begin{algorithm}[!ht]
\caption{Prior Performance via Distill-and-Evaluate}
\label{alg:prior_perf}
\small
\begin{algorithmic}[1]
\Require Committee $\mathcal{S}$, dataset $\mathcal{D}$
\State Split $\mathcal{D}$ into $\mathcal{D}_{\mathrm{tr}}$ (80\%) and $\mathcal{D}_{\mathrm{ev}}$ (20\%).
\For{$\Phi \in \mathcal{S}$}
  \State Train $\Phi$ on $\mathcal{D}_{\mathrm{tr}}$ with $\mathcal{L}_{\mathrm{CE}}$; distill $\mathcal{D}^{\mathrm{dist}}_{\Phi}$ using $\Phi$.
  \State Train student $f_{\Phi}$ on $\mathcal{D}^{\mathrm{dist}}_{\Phi}$ with $\mathcal{L}_{\mathrm{KL}}=\mathrm{KL}(\Phi(x)\,\|\,f_{\Phi}(x))$; set $\alpha_{\Phi}\!\gets\!\text{Acc}(f_{\Phi},\mathcal{D}_{\mathrm{ev}})$.
\EndFor
\State \Return $\{\alpha_{\Phi}\}_{\Phi\in\mathcal{S}}$
\end{algorithmic}
\end{algorithm}

\subsection{Overview of \name{}}
\label{sec:method}
The overall framework of \name{} is shown in Fig.~\ref{fig:overview}. Built upon the enhanced baseline SRe$^2$L++, \name{} introduces a Prior Performance Guided Voting Strategy that assigns greater influence to stronger models, addressing limitations of prior ensemble methods and improving effectiveness.

\subsection{Prior Performance Guided Voting Strategy}
\begin{theorem}[Proof in Appendix~\ref{proof:committee_diversity}]
\label{thm:diversity-separation}
Let \( \{\Phi_i\}_{i=1}^{|S|} \) be a committee of models, with diversity quantified as \( K := \frac{2}{|S|(|S|-1)}\sum_{i<j} \Pr_{x \sim \mathcal{D}}[\Phi_i(x) \ne \Phi_j(x)] \). 
We denote the gradient at iteration \( t \) for sample $\hat{u}_z$ as \( G_z^{(t)} \), and suppose:
(i) \( \|G_z^{(t)}\|^2 \le G_{\max} \),
(ii) \( \mathbb{E}[\|G_z^{(t)} - G_{z'}^{(t)}\|^2] \ge C_g K \) for \( z, z' \) from the same class and some constant $C_g$.
Then the expected cosine distance satisfies:
\begin{equation}
\mathbb{E}[\Delta_{\cos}^{(t+1)}] \ge \mathbb{E}[\Delta_{\cos}^{(t)}] + \frac{1}{2}\eta^2 C_g K,
\end{equation}
where \( \Delta_{\cos}^{(t)} := 1 - \cos(\hat{u}_z^{(t)}, \hat{u}_{z'}^{(t)}) \) quantifies the discrepancy between normalized update directions. Thus, greater committee diversity yields increased intra-class separation.
\end{theorem}

\begin{figure}
    \centering
    \includegraphics[width=0.93\linewidth]{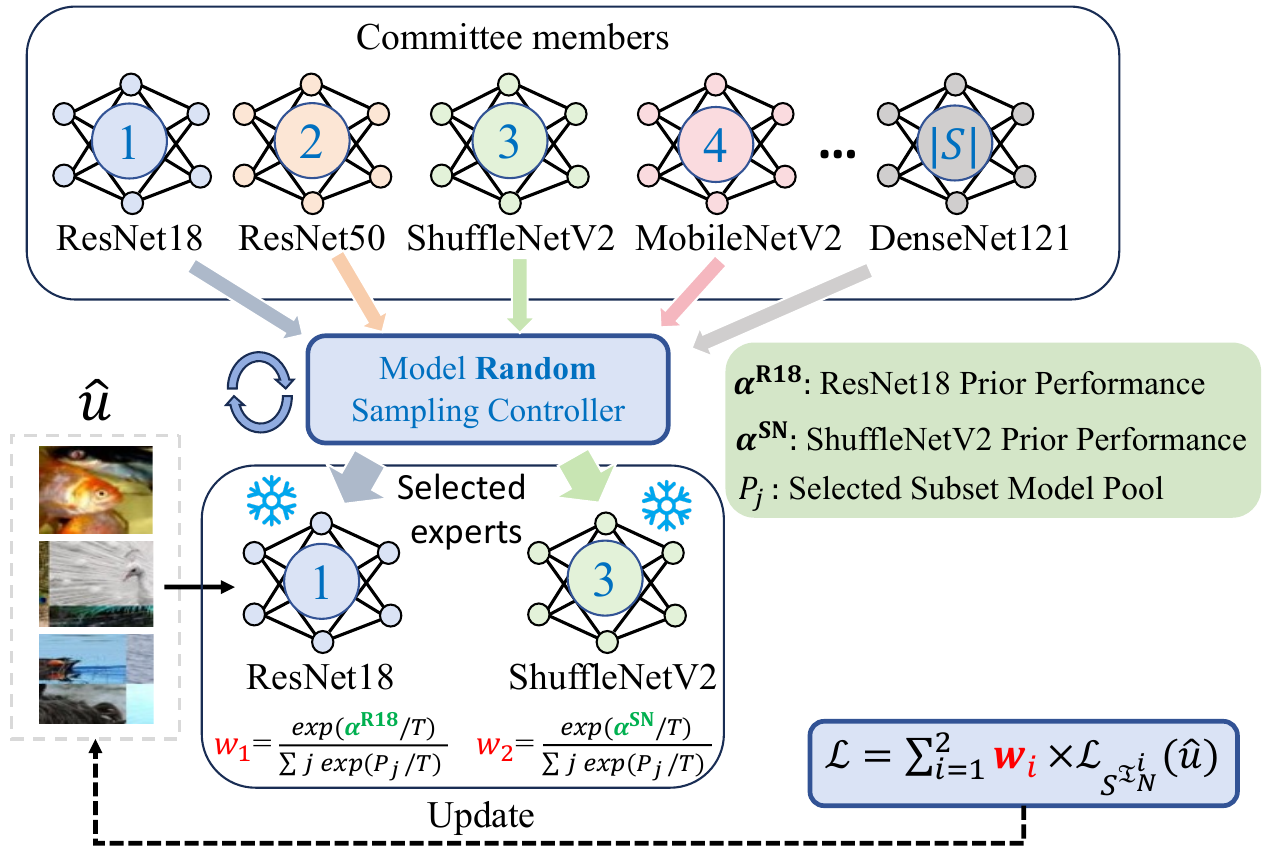}
    \caption{Demonstration of our proposed {\em Committee Voting}.}
    \label{fig:detail_vis_cv}
    \vspace{-.2in}
\end{figure}

\textbf{Committee Members.} Let \(S\) denote the committee comprising \(|S|\) diverse backbone architectures, which are selected to introduce architectural diversity, enrich representation capacity, and improve the robustness of the distilled dataset (see Theorem~\ref{thm:diversity-separation} and right part of Fig.~\ref{fig:performance_and_cos_sim}).

\noindent\textbf{Prior Performance.} We generate distilled data from pre-trained models, with its quality reflecting the critical information retained. Thus, the generalization of models trained on this data serves as a proxy for the source model's prior performance (detailed in Algorithm~\ref{alg:prior_perf}).

\noindent\textbf{Prior-based Voting in Distilled Data Generation.} See Fig.~\ref{fig:detail_vis_cv} for details. Let \(\mathcal{I}_N \subset \{1, \dots, |S|\}\) be a randomly sampled subset of \(N\) indices, where $2 \leq N \leq |S|$. The \(i\)-th index in \(\mathcal{I}_N\) is denoted by \(\mathcal{I}_{N}^{i}\), with the corresponding backbone and its prior performance represented as \(S^{\mathcal{I}_{N}^{i}}\) and \(\alpha^{\mathcal{I}_{N}^{i}}\), respectively. We define the prior-based voting loss as:
\begin{align}
\mathcal{L}(\hat{u}) &= \sum_{i=1}^{N} \frac{\exp(\alpha^{\mathcal{I}_{N}^{i}}/T)}{\sum_{j=1}^{N} \exp(\alpha^{\mathcal{I}_{N}^{j}}/T)} \, \mathcal{L}_{S^{\mathcal{I}_{N}^{i}}}(\hat{u}),
\label{DD-CV equation}
\end{align}
where SoftMax prioritizes stronger models, steering optimization toward informative directions. In practice, using a small subset (e.g., N=2) balances the influence of the 
strongest expert while retaining complementary signals.

\noindent\textbf{Prior-based Voting in Soft Label Generation.} Similarly, soft labels are generated through prior-guided voting, where class probabilities from selected models are aggregated via a weighted average determined by their prior performance. This strategy amplifies the influence of informative models, leading to more reliable supervision.

\begin{theorem}[Proof in Appendix~\ref{proof:ppg is better}]
\label{thrm:ppg is better}
Let $J(\hat{u})$ denote the generalization risk of the synthetic image. Suppose each model $\Phi_i$ is associated with a prior performance score $\alpha_i > 0$, and contributes a gradient $\nabla_{\hat{u}} \ell(\Phi_i(\hat{u}), \hat{v})$ during optimization. Assume that $\left\langle \nabla J(\hat{u}), \nabla_{\hat{u}} \ell(\phi_i(\hat{u}), \hat{v}) \right\rangle = \lambda \cdot \alpha_i, \quad \lambda > 0.$ Then, the inner product between the generalization risk gradient and the aggregate update under prior-weighted voting satisfies:
\begin{equation}
   \left\langle \nabla J(\hat{u}), G_{\text{prior}} \right\rangle > \left\langle \nabla J(\hat{u}), G_{\text{equal}} \right\rangle, 
\end{equation}
where $G_{\text{prior}}$ and $G_{\text{equal}}$ are weighted and uniform gradient averages, with each weight given by a SoftMax over $\alpha_i$.
\end{theorem}
The key insight is that prior-guided voting aligns updates more closely with the gradient direction that promotes generalization, compared to uniform averaging.

\begin{figure}[t]
    \centering
    \includegraphics[width=1\linewidth]{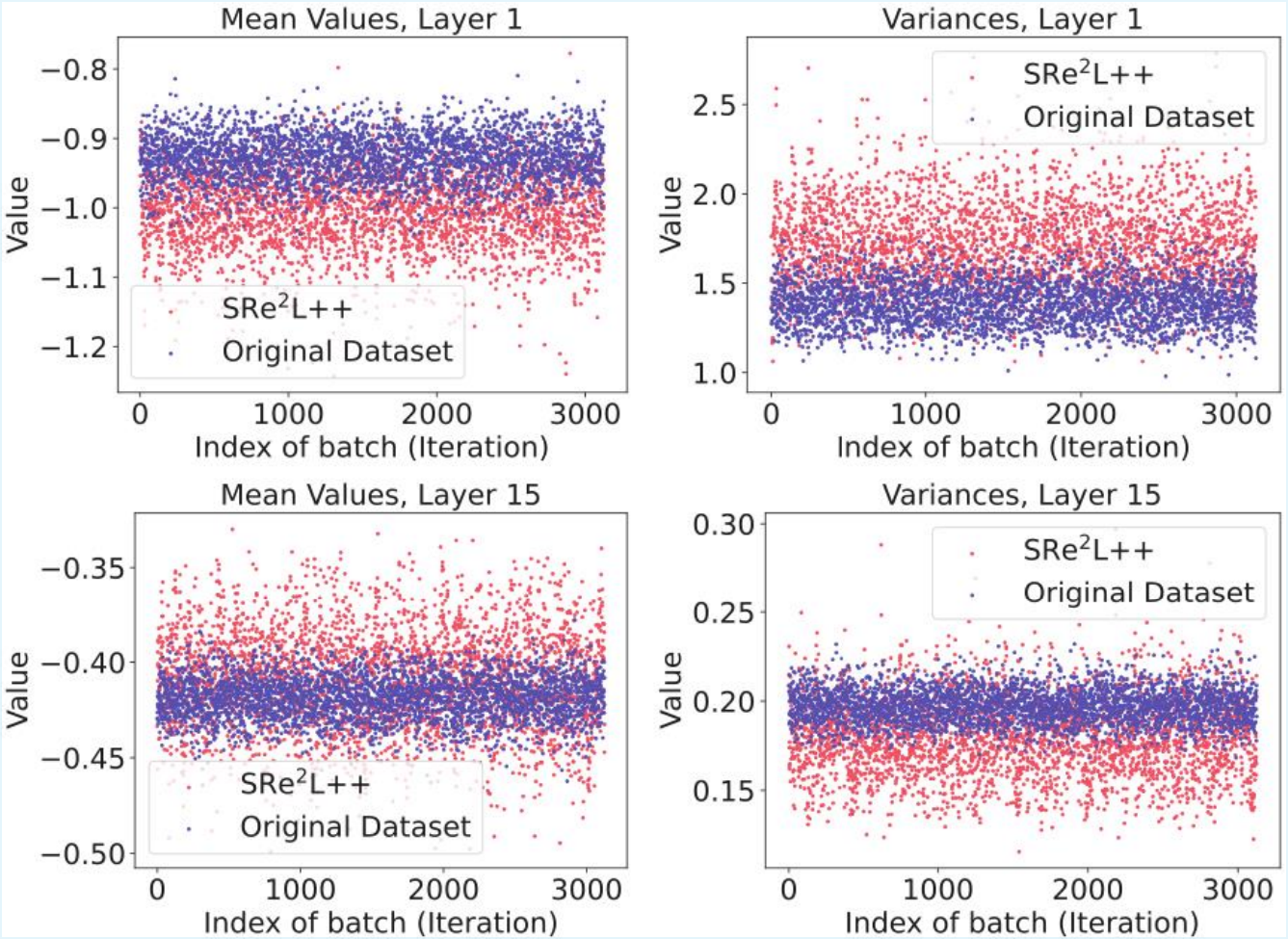}
    \vspace{-1em}
    \caption{Feature-level statistical discrepancies between synthetic data generated by SRe$^2$L++ and the training data on ImageNet-1K, evaluated across different batches in a pre-trained ResNet18.}
    \label{fig:bn_bias}
    \vspace{-.2in}
\end{figure}

\subsection{Batch-Specific Soft Labeling}
\label{bssl}
\begin{figure}[!t]
    \centering
    \includegraphics[width=1\linewidth]{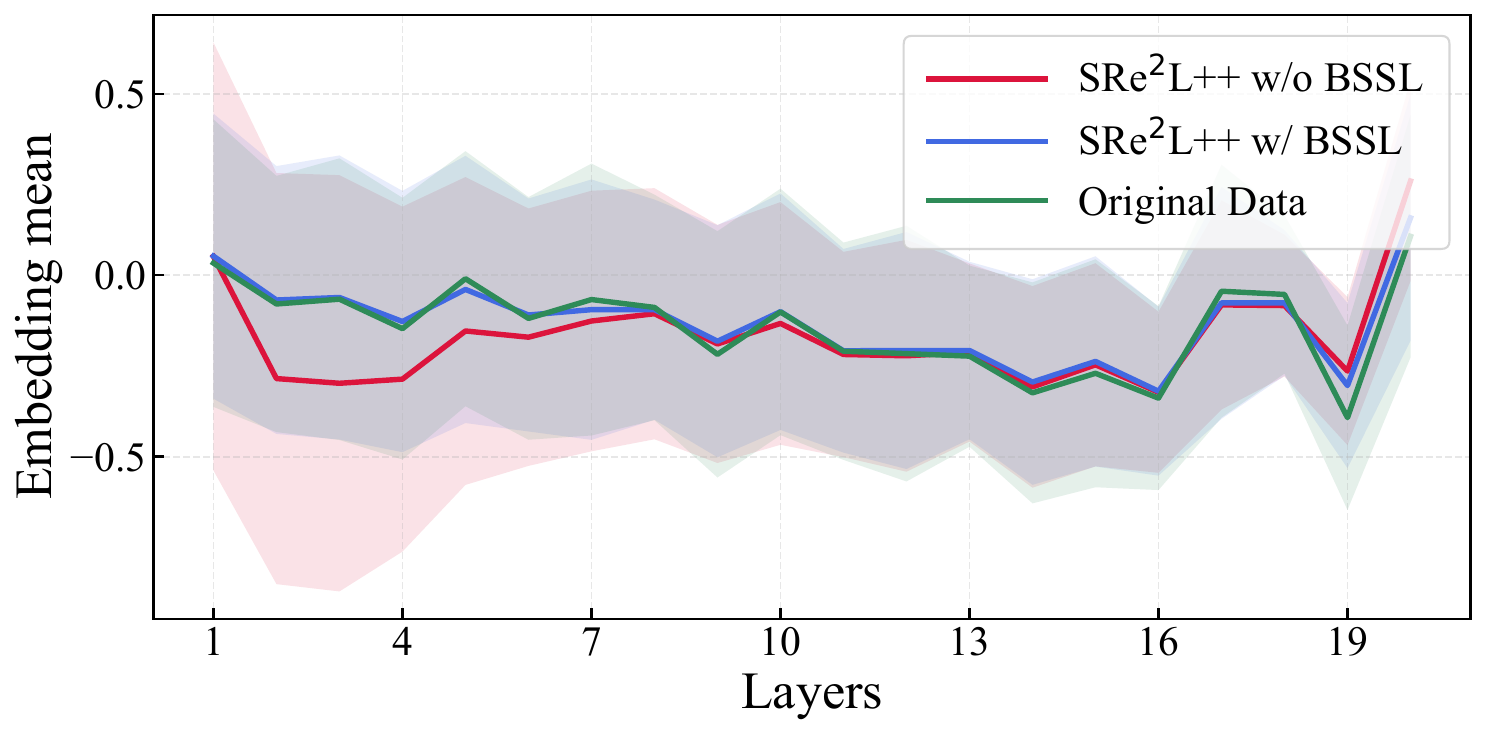}
    \vspace{-.25in}
    \caption{Embedding statistics w/ and w/o BSSL across layers.}
    \label{fig:bssl_effect}
    \vspace{-.2in}
\end{figure}

\noindent In the post-evaluation stage, a teacher model is commonly employed to pre-generate soft labels~\citep{shen2022fast}, thereby enhancing the generalization of the student model~\citep{hinton2015distilling,soft_label}. Typically, the teacher model includes Batch Normalization layers~\citep{RDED_2024,GBVSM_2024,sre2l_2024,qin2024label}, which utilize running statistics to normalize features. These statistics are progressively updated during training,
\begin{align*}
\mu_{\text{running}} &\leftarrow \lambda \, \mu_{\text{running}} + (1 - \lambda) \, \mu_{B}, \\
\sigma^2_{\text{running}} &\leftarrow \lambda \, \sigma^2_{\text{running}} + (1 - \lambda) \, \sigma^2_{B}.
\end{align*}
where \( \lambda \) denotes the momentum, and \( \mu_B \) and \( \sigma^2_B \) are the mean and variance of the current batch, respectively. As shown in Fig.~\ref{fig:bn_bias}, even when synthesized images are optimized to match BN statistics during generation, a notable gap remains between the BN distributions of synthetic and real data due to regularization effects and optimization randomness. To address this issue, we propose {\em Batch-Specific Soft Labeling} (BSSL). Instead of using pre-trained BN statistics from real data, BSSL recomputes BN statistics from each synthetic batch while keeping all other teacher parameters fixed when generating soft labels. This simple modification substantially improves post-training performance on synthetic data.
Given a distilled batch \( B = \{\hat{u}_i\}_{i=1}^N \), where \( \hat{u}_i \in \mathbb{R}^{C \times H \times W} \) denotes the feature map of sample \( i \), the BatchNorm statistics are computed as:
\begin{align}
\mu_{B,c} &= \frac{1}{NHW} \sum_{i,h,w} \hat{u}_{i,c,h,w}, \\
\sigma^2_{B,c} &= \frac{1}{NHW} \sum_{i,h,w} (\hat{u}_{i,c,h,w} - \mu_{B,c})^2 + \epsilon
\end{align}
where \( \mu_{B, c} \) and \( \sigma^2_{B, c} \) denote the batch statistics for channel \( c \), with a small \( \epsilon \) added for numerical stability. This refinement aligns normalization statistics more closely, improving soft-label quality. As shown in Fig.~\ref{fig:bssl_effect}, BSSL yields more consistent embedding statistics across BN layers.

\textbf{Extending BSSL to Non-BN Architectures.} For models without native BatchNorm, BSSL can be enabled by explicitly introducing BatchNorm layers. As a concrete instantiation, we adopt a BN-ViT variant that replaces LayerNorm with BatchNorm and inserts BN layers into feed-forward blocks~\citep{yao2021leveraging}. This demonstrates that BSSL does not fundamentally rely on BN-native designs.

\begin{table*}[!htp]
    \centering
    \caption{\textbf{Comparison with SOTA Baseline Methods.}  All models are trained with 300 epochs.} 
    \vspace{-.5em}
    \label{main_table}
    \resizebox{0.96\textwidth}{!}{
    \begin{tabular}{cc|cccc|cccc|cccc}
    \toprule
    & &  \multicolumn{4}{c|}{ResNet18} & \multicolumn{4}{c|}{ResNet50} & \multicolumn{4}{c}{ResNet101} \\ 
    \cmidrule(lr){3-6} 
    \cmidrule(lr){7-10} 
    \cmidrule(lr){11-14} 
    Dataset & IPC (Ratio) & CDA & RDED & SRe\(^2\)L++ & \name{} & CDA & RDED & SRe\(^2\)L++ & \name{} & CDA & RDED & SRe\(^2\)L++ & \name{} \\
    \midrule
    \multirow{3}{*}{CIFAR-10} 
    & 1 (0.02\%) & - & 22.9 & 24.9 & \textbf{27.4} & - & 10.2 & 24.4& \textbf{24.9} & - & 18.7 & 23.4 & \textbf{26.6} \\
    & 10 (0.2\%)& - & 37.1 & 51.3 & \textbf{54.7} & - & 33.1 & 47.9 & \textbf{49.7} & - & 33.7 &  41.5 & \textbf{48.2} \\
    & 50 (1.0\%)& - & 62.1 & 75.8 & \textbf{76.9} & - & 54.2 & 71.4& \textbf{72.1} & - & 51.6 &  73.6& \textbf{74.4} \\
    \midrule
    \multirow{3}{*}{CIFAR-100}
    & 1 (0.2\%) & 13.4 & 11.0 & 12.0 & \textbf{16.5} & - & 10.9 & 12.6 & \textbf{17.8} & - & 10.8 & 9.9 & \textbf{15.1} \\
    & 10 (2.0\%) & 49.8 & 42.6 &  56.7& \textbf{61.8} & - & 41.6 & 52.1 & \textbf{59.6} & - & 41.1 & 54.3 & \textbf{61.5} \\
    & 50 (10.0\%) &64.4 & 62.6 & 66.6 & \textbf{69.4} & - & 64.0 & 67.0 & \textbf{70.3} & - & 63.4 & 67.8  & \textbf{71.1} \\
    \midrule
    \multirow{3}{*}{Tiny-ImageNet}
    & 1 (0.2\%) &3.3 & 9.7 & 9.3 & \textbf{20.1} & - & 8.2 & 7.6 & \textbf{17.5} & - & 3.8 & 7.5 & \textbf{19.2} \\
    & 10 (2.0\%) &43.0 & 41.9 & 46.5 & \textbf{53.0} & - & 38.4 & 42.8 & \textbf{52.8} & - & 22.9 & 45.4 & \textbf{53.9} \\
    & 50 (10.0\%) &48.7 & 58.2 & 53.5 & \textbf{61.2} & 49.7 & 45.6 & 53.7 & \textbf{64.1} & 50.6 & 41.2 & 53.7 & \textbf{63.1} \\
    \midrule
    \multirow{3}{*}{ImageNette}
    & 1 (0.1\%)& - & 35.8 & 34.7 & \textbf{37.5} & - & 27.0 & 25.9 & \textbf{29.4} & - & 25.1 & 25.2  &  \textbf{26.9}\\
    & 10 (1.0\%)& - & 61.4 & 73.7 & \textbf{74.4} & - & 55.0 & 72.6 & \textbf{73.9} & - & 54.0 & 66.6 & \textbf{67.8} \\
    & 50 (5.2\%)&- & 80.4 & 80.3 & \textbf{81.4} & - & 81.8 & 81.2 & \textbf{82.3} & - &75.0  &73.4 & \textbf{76.3} \\
    \midrule
    \multirow{3}{*}{ImageWoof}
    & 1 (0.1\%) &- &20.8  & 17.0 &\textbf{21.3}& - & 17.8  & 14.7 &\textbf{22.2}& - &19.6 &12.9  &\textbf{20.4}\\
    & 10 (1.1\%)&- &38.5 & 45.2 & \textbf{63.0} & - & 35.2 & 42.3 & \textbf{47.5}& - & 31.3 & 38.3 & \textbf{45.2}\\
    & 50 (5.3\%)&- & 68.5 & 64.5 & \textbf{68.7}& - & 64.1 & 65.5 & \textbf{68.7} & - & 59.1 &  57.8 &\textbf{59.6} \\
    \midrule
    \multirow{3}{*}{ImageNet-1k}
    & 1 (0.1\%) &- & 6.6 & 8.6 & \textbf{12.1} & - & 8.0 & 8.0 & \textbf{12.8} & - & 5.9 & 6.2 & \textbf{8.4} \\
    & 10 (0.8\%) & 33.6 & 42.0 & 43.1 & \textbf{49.5} & - & 49.7 & 47.3 & \textbf{57.0} & - & 48.3 & 51.2 & \textbf{57.2} \\
    & 50 (3.9\%) & 53.5 & 56.5 & 57.6 & \textbf{59.5} & 61.3 & 62.0 & 61.8 & \textbf{65.3} & 61.6 & 61.2 & 61.0 & \textbf{64.6} \\
    \bottomrule
    \end{tabular}
    }
    \vspace{-0.1in}
\end{table*}

\section{Experiments}
\label{sec:exp}

\textbf{Datasets.} To broadly evaluate the performance of \name{}, we conduct experiments on low- and high-resolution datasets. The low-resolution setting includes CIFAR-10/100~\citep{cifar10}, while the high-resolution setting covers Tiny-ImageNet~\citep{le2015tiny}, ImageNet-1K~\citep{deng2009imagenet}, and its subsets.

\textbf{Baseline Methods.} {To evaluate the effectiveness of our tailored ensemble strategy, we include three ensemble-based methods: MTT, G-VBSM, and EDC.} Additionally, we consider CDA, RDED and SRe$^2$L++, which incorporate recent advances in dataset distillation.

\subsection{Main Results}

\textbf{High- and Low-Resolution Datasets.} As shown in Table~\ref{main_table}, \name{} consistently outperforms state-of-the-art distillation methods across large- and small-scale benchmarks. On ImageNet-1K at IPC=50 with ResNet-18, it achieves 59.5\%, exceeding SRe$^2$L++ by +1.9\%. On CIFAR-100 at IPC=10, \name{} reaches 61.8\%, outperforming RDED, SRe$^2$L++, and CDA by +19.2\%, +5.1\%, and +12\%, respectively. These results demonstrate the robustness and broad applicability of \name{} across datasets of varying scales and resolutions.

\textbf{Comparison with Vanilla Ensemble Methods.} To ensure fair comparison with EDC, \name{} is evaluated under the same settings. As shown in Table~\ref{tab:ensemble_comp}, \name{} consistently outperforms prior vanilla ensemble methods across datasets and resolutions. Notably, at IPC=50, it exceeds EDC by +1.5\% on ImageNet-1K and MTT by +33.2\% on Tiny-ImageNet, validating the effectiveness of \name{}'s tailored ensemble strategy for high-quality dataset distillation.

\begin{table}[!t]
    \caption{Comparison between vanilla ensemble methods and our prior-based voting \name{}, using ResNet18 for G-VBSM, EDC, and ours, and Conv128 for MTT on long training setting.}
    \vspace{-.05in}
    \label{tab:ensemble_comp}
    \centering
    \renewcommand{\arraystretch}{1}
    \resizebox{0.88\linewidth}{!}{  
        \begin{tabular}{cc|cccc}
            \toprule
            Dataset &  IPC & MTT & G-VBSM & EDC & \name{}\\
            \midrule
            \multirow{2}[0]{*}{CIFAR-100} 
              & 10 & 40.1  & 59.5  & 63.7& \textbf{66.0} \\
              & 50 & 47.7 & 65.0 & 68.6&  \textbf{70.4} \\
            \midrule
            \multirow{2}[0]{*}{Tiny-ImageNet} 
              & 10 & 23.2  & - & 51.2 &\textbf{53.0} \\
              & 50 & 28.0  & 47.6 &57.2 & \textbf{61.2} \\
            \midrule
            \multirow{2}[0]{*}{ImageNet-1k}    
              & 10 & - & 31.5 & 48.6 &\textbf{49.5}  \\
              & 50 & - & 51.8  &58.0& \textbf{59.5} \\
            \bottomrule
        \end{tabular}
    }
    \vspace{-.1in}
\end{table}

\subsection{Ablation Study}
\label{Ablation}

\textbf{Impact of \( T \).} As shown in Table~\ref{tab:different_T}, \( T=5 \) achieves the best overall performance. A low temperature (\( T=1 \)) overly amplifies score differences and leads to model dominance, whereas a high temperature (\( T=20 \)) yields near-uniform expert weighting, degrading distilled data quality.

\textbf{Impact of Prior-Based Voting.} To assess the effectiveness of prior voting, we conduct ablation studies in Table~\ref{tab:prior_effect}. The prior voter consistently outperforms equal and random voters across all settings, demonstrating the benefit of leveraging prior knowledge for high-quality dataset generation.

\begin{table*}[t]
    \centering
    \small
    \caption{Ablation experiments. For (a) to (c), The Top row is conducted on CIFAR-100; while the Bottom row is conducted on ImageNet-1K. All results are reported with IPC=10 using ResNet-18.
    } 
    \vspace{-.05in}
    \begin{subtable}{0.32\linewidth}
        \centering
        \resizebox{1\linewidth}{!}{ 
        \begin{tabular}{ccccc}
        \toprule
         $T=1$  & $T=5$ & $T=10$ & $T=15$ & $T=20$ \\
        \midrule
          59.3  & \textbf{61.8} & 61.3 &  61.0 & 60.9 \\
         45.8  &  \textbf{49.5} &  49.2 &   49.0 & 48.8 \\
        \bottomrule
        \end{tabular}}
        \caption{Performance across different temperature values ($T$).}
        \label{tab:different_T}
    \end{subtable}
    \hfill
    \begin{subtable}{0.5\linewidth}
            \centering
            \resizebox{1\linewidth}{!}{ 
            \begin{tabular}{cccc}
                \toprule
                \cjc{SRe$^2$L++} & CV-DD w/ Random & CV-DD w/ Equal & CV-DD w/ Prior \\
                \midrule
                 \cjc{56.7} &59.8 & 60.7 &  \textbf{61.8} \\
                \cjc{43.1} &47.6 & 48.2 &  \textbf{49.5} \\
                \bottomrule
            \end{tabular}}
            \caption{ Comparison of weighting strategies, where \textit{Equal} assigns uniform weights (0.5) and \textit{Random} samples weights arbitrarily.}
            \label{tab:prior_effect}
    \end{subtable}
    \hfill
    \begin{subtable}{0.13\linewidth}
            \centering
             \resizebox{1\linewidth}{!}{ 
            \begin{tabular}{cc}
                \toprule
                $N=2$  & $N=3$ \\
                \midrule
                \textbf{61.8}  &  60.1 \\
                \textbf{49.5}  &  48.7 \\
                \bottomrule
            \end{tabular}}
            \caption{Effect of numbers of experts.}
            \label{tab:experts_num}
    \end{subtable}

    \begin{subtable}{0.47\linewidth}
        \centering
        \resizebox{1\linewidth}{!}{ 
        \begin{tabular}{ccccc|cc}
            \toprule
            \multicolumn{5}{c|}{\textbf{Committee Choices} $S$} & \multicolumn{2}{c}{\textbf{Student Accuracy (\%)}} \\
            \cmidrule(lr){1-5} \cmidrule(lr){6-7}
            R18 & R50 & D121 & SV2 & MBV2 & CIFAR-100 & ImageNet-1K \\
            \midrule
            \checkmark &  &   &  &  & 56.7 & 43.1 \\
            \checkmark & \checkmark &  &  &  & 60.0 & 43.8 \\
            \checkmark & \checkmark & \checkmark &  &  & 61.0 & 45.4 \\
            \checkmark & \checkmark & \checkmark & \checkmark &   & 61.5 & 48.3 \\
            \checkmark & \checkmark & \checkmark & \checkmark & \checkmark & 61.8 & 49.5 \\
            \bottomrule
        \end{tabular}}
        \caption{Effect of committee size ($S$) on the performance of the student model (ResNet-18) trained with a 10-IPC distilled dataset. }
        \label{tab:committe-size-ablation}
    \end{subtable}
    \hfill
    \begin{subtable}{0.51\linewidth}
            \centering
            \resizebox{1\linewidth}{!}{ 
           \begin{tabular}{lccccc}
            \toprule
            \multirow{2}{*}{Method} & \multirow{2}{*}{IPC} 
            & \multicolumn{2}{c}{CIFAR-100} 
            & \multicolumn{2}{c}{ImageNet-1k} \\
            \cmidrule(lr){3-4} \cmidrule(lr){5-6}
            & & w/o BSSL & w/ BSSL & w/o BSSL & w/ BSSL \\
            \midrule
            \multirow{2}{*}{SRe$^2$L++} 
            & 1  & 10.5 & \textbf{12.0} & 4.2  & \textbf{8.6} \\
            & 10 & 53.3  & \textbf{56.7} & 38.5  & \textbf{43.1} \\
            \midrule
            \multirow{2}{*}{\name{}} 
            & 1  & 12.9 & \textbf{16.5} & 5.8  & \textbf{12.1} \\
            & 10 & 57.3  & \textbf{61.8} & 42.5 & \textbf{49.5} \\
            \bottomrule
            \end{tabular}
            }
        \caption{Performance comparison on CIFAR-100 and ImageNet-1K under varying IPCs, with and without BSSL. }
        \label{tab:BSSL-ablation}
    \end{subtable}
    \vspace{-2em}
\end{table*}

\textbf{Impact of the Number of Experts \( N \).}  
As shown in Table~\ref{tab:experts_num}, increasing \(N\) to 3 increases computational cost and degrades performance by diluting the strongest model’s influence, which weakens gradient alignment and reduces dataset fidelity. Thus, \(N = 2\) offers the best trade-off.

\textbf{Impact of Committee Size $|S|$.} Table~\ref{tab:committe-size-ablation} shows that larger committees consistently improve distilled dataset quality, highlighting the benefit of diverse expertise.

\textbf{Impact of BSSL.} Table~\ref{tab:BSSL-ablation} shows that BSSL improves performance under synthetic--real distribution shift, yielding a +7.0\% gain for \name{} at IPC=10 on ImageNet-1K.

\subsection{Analysis}

\begin{figure}[t]
    \centering
    \vspace{-0.1in}
    \includegraphics[width=0.7\linewidth]{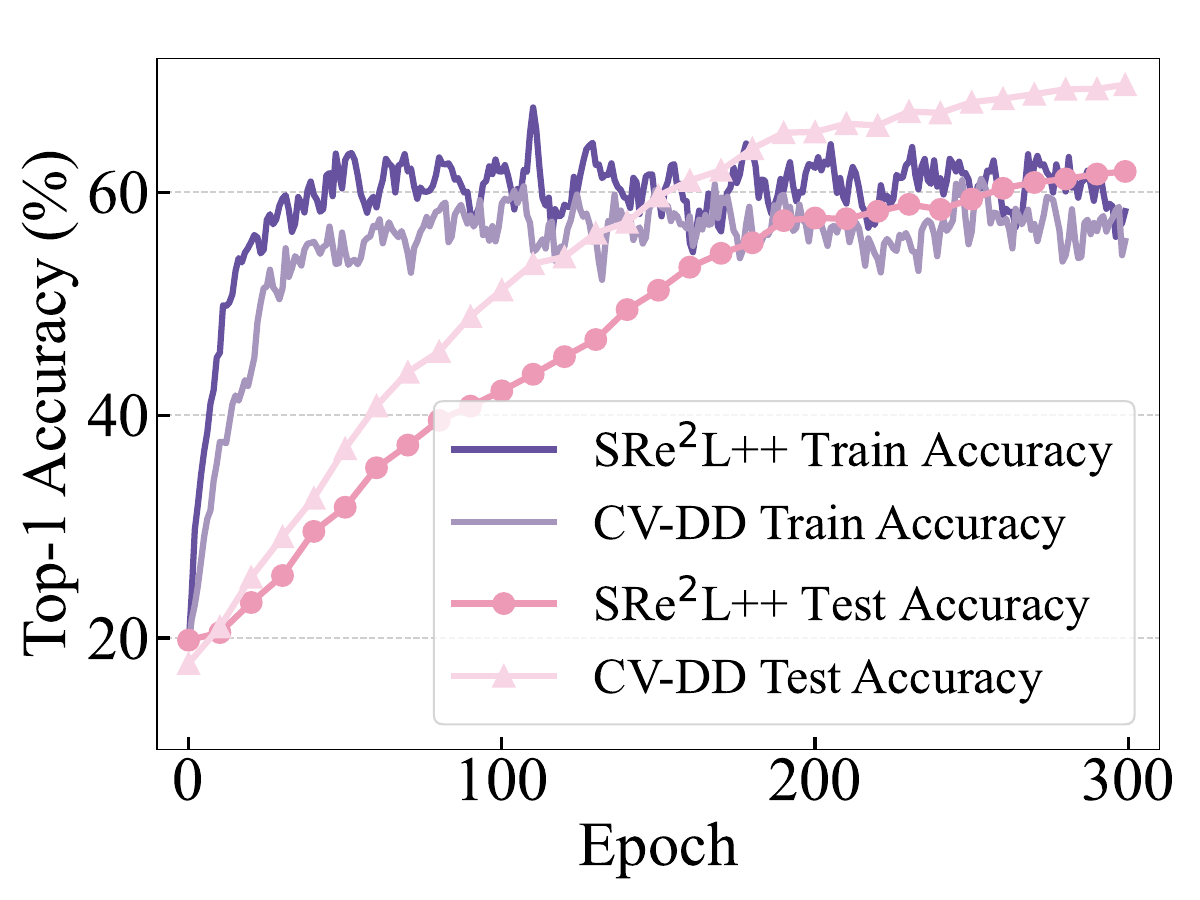}
    \vspace{-.1in}
    \caption{Comparison of Top-1 accuracy curve between \name{} and SRe$^2$L++ on CIFAR-10 with 50 IPC.}
    \label{fig:overfitting_effect_bssl}
    \vspace{-.15in}
\end{figure}

\textbf{Overfitting Analysis.} \name{} effectively mitigates overfitting during post-training. As shown in Fig.~\ref{fig:overfitting_effect_bssl}, it achieves lower training accuracy but higher test accuracy than SRe$^2$L++, indicating that \name{}’s Prior Voting acts as an effective regularizer in overfitting-prone scenarios.

\begin{wraptable}{r}{0.3\columnwidth}
\vspace{-.18in}
\caption{Impact of overfitted teacher.}
\vspace{-0.05in}
\label{tab:overfitted-teacher}
\hspace{-0.2in}
\vspace{-.05in}
\resizebox{1.15\linewidth}{!}{
    \begin{tabular}{cc}
        \toprule
        \cjc{Standard} &  \cjc{Overfitted} \\
        \midrule
         \cjc{54.7} & \cjc{54.3} \\
        \bottomrule
    \end{tabular}
}
\vspace{-.1in}
\end{wraptable}
\textbf{Robustness to Overfitted Teachers.} 
Overfitted teachers introduce unstable signals that can harm distillation. CV-DD mitigates this through prior voting, which inherently assigns them small weights. As shown in Table~\ref{tab:overfitted-teacher}, replacing a standard MobileNetV2 with an overfitted one yields only a minor drop, confirming CV-DD’s robustness to noisy teachers.

\textbf{Cross-Framework Generalization.} The committee mechanism integrates seamlessly into diverse dataset distillation frameworks. We validate this by applying it to the non--training-based RDED, where committee voting yields consistent gains (Table~\ref{tab:rdded_cvdd}). This shows CV-DD as a versatile plug-in driven by model diversity and prior-guided voting.

\begin{table}[!h]
\caption{Comparison of RDED, equal voting, and prior voting.}
\label{tab:rdded_cvdd}
\vspace{-.05in}
\centering
\resizebox{0.7\linewidth}{!}{
\begin{tabular}{cccc}
\toprule
& RDED & w/ Equal Voting &  w/ Prior Voting  \\
\midrule
IPC = 10 & \cjc{42.0} & \cjc{43.2} & \cjc{\textbf{44.8}} \\
IPC = 20& \cjc{47.9} & \cjc{48.5} & \cjc{\textbf{49.7}} \\
IPC = 30 & \cjc{51.7} & \cjc{52.1} & \cjc{\textbf{53.2}} \\
\bottomrule
\end{tabular}
}
\vspace{-.1in}
\end{table}

\textbf{Cross-Architecture Generalization.} We evaluate \name{} against RDED, EDC, and SRe$^2$L++ across nine diverse architectures. As shown in Table~\ref{tab:cross-arch}, \name{} achieves the best accuracy across all architectures, demonstrating robust and consistent cross-model generalization.

\begin{table}[!ht]
    \vspace{-0.05in}
    \caption{Top-1 accuracy on ImageNet-1K for cross-architecture generalization with IPC=10.}
    \vspace{-.05in}
    \label{tab:cross-arch}
    \centering
    \renewcommand{\arraystretch}{1}
    \resizebox{0.8\linewidth}{!}{  
        \begin{tabular}{lccccc}
            \toprule
            Model & \#Params & RDED & EDC & SRe$^2$L++ & \name{}  \\
            \midrule
            ShuffleNetV2 & 1.4M & 19.6 & 29.8 & 22.9 & \textbf{30.6}  \\
            MobileNetV2 & 3.4M  & 34.4 & 45.0 & 37.1 & \textbf{45.6}  \\
            DenseNet121 & 8.0M & 49.4 & - & 46.7 & \textbf{54.7}  \\
            ResNet18 & 11.7M  & 42.0 & 48.6 & 43.1 & \textbf{49.5}  \\
            ResNet50 & 25.6M  & 49.7 & 54.1 & 47.3 & \textbf{57.0}  \\
            Swin-Tiny & 28.0M & 29.2 & 38.3 & 28.3 & \textbf{39.2} \\
            ResNet101 & 44.5M & 48.3 & 51.7 & 51.2 & \textbf{57.2}  \\
            RegNetX-8gf & 39.6M & 51.9 & - & 53.4 & \textbf{60.9}  \\
            WRN-50-2 & 68.9M & 50.0 & - & 50.2 & \textbf{58.3}  \\
            \bottomrule
        \end{tabular} 
    }
    \vspace{-.1in}
\end{table}

\begin{wraptable}{r}{0.4\columnwidth}
\vspace{-.18in}
\caption{Performance w/ and w/o BSSL.}
\vspace{-.05in}
\label{tab:vit-bssl}
\centering
    \resizebox{1\linewidth}{!}{ 
\begin{tabular}{cc}
\toprule
\cjc{w/o BSSL} & \cjc{w/ BSSL} \\
\midrule
\cjc{16.5} & \cjc{18.4} \\
\toprule
\end{tabular}
}
\vspace{-.18in}
\end{wraptable}
\textbf{Incorporating ViT in BSSL.}
To evaluate the BN-ViT adaptation, we compare BN-ViT with and without BSSL. As shown in Table~\ref{tab:vit-bssl}, BN-ViT+BSSL achieves higher accuracy, indicating that BSSL extends effectively to ViT architectures. This suggests that BN reliance does not limit the applicability of BSSL beyond BN-native models.

\begin{wraptable}{r}{0.33\columnwidth}
\vspace{-.16in}
\caption{Synthetic-to-real transfer.}
\label{tab:cross-task}
\vspace{-.05in}
\centering
\resizebox{1\linewidth}{!}{
    \begin{tabular}{cc}
        \toprule
         SRe$^2$L++ & CV-DD \\
        \midrule
     18.9 & \textbf{20.7} \\
        \bottomrule
    \end{tabular}
}
\vspace{-.1in}
\end{wraptable}
\textbf{Synthetic-to-Real Transfer Tasks.}
To assess generalization beyond standard classification, we evaluate CV-DD on the synthetic-to-real VisDA-2017 benchmark~\citep{peng2017visda}. Using the same pipeline, we distill IPC=10 datasets with CV-DD and SRe$^2$L++, train classifiers, and evaluate on the real-domain validation set. As shown in Table~\ref{tab:cross-task}, CV-DD surpasses SRe$^2$L++ by +1.8\%, demonstrating robustness under substantial distribution shift.

\begin{table*}[!t]
    \centering
    \small
    \caption{\cjc{Detailed prior-evaluation time and total runtime (including prior evaluation) for distillation.}}
    \label{tab:prior-evaluation}

    \begin{subtable}{0.6\linewidth}
        \centering
        \resizebox{1\linewidth}{!}{%
        \begin{tabular}{lccccc}
        \toprule
        \cjc{\textbf{Model}} & \cjc{\textbf{Pretraining}} & \cjc{\textbf{Generation}} & \cjc{\textbf{Evaluation}} & \cjc{\textbf{Peak GPU Usage}} & \cjc{\textbf{Total Time}} \\
        \midrule
        \cjc{ResNet-18} & \cjc{10.1 h} & \cjc{1.3 h} & \cjc{1.5 h} & \cjc{11.3 GB} & \cjc{12.9 h} \\
        \cjc{ResNet-50} & \cjc{14.4 h} & \cjc{4.4 h} & \cjc{1.5 h} & \cjc{23.0 GB} & \cjc{20.3 h} \\
        \cjc{ShuffleNetV2} & \cjc{9.4 h} & \cjc{2.6 h} & \cjc{1.5 h} & \cjc{7.6 GB} & \cjc{13.4 h} \\
        \cjc{MobileNet-V2} & \cjc{10.2 h} & \cjc{2.4 h} & \cjc{1.5 h} & \cjc{11.4 GB} & \cjc{14.1 h} \\
        \cjc{DenseNet-121} & \cjc{16.4 h} & \cjc{6.0 h} & \cjc{1.5 h} & \cjc{19.5 GB} & \cjc{23.9 h} \\
        \bottomrule
        \end{tabular}}
        \caption{\cjc{Detailed prior evaluation time.}}
        \label{tab:prior-evaluation-time}
    \end{subtable}
    \hfill
    \begin{subtable}{0.33\linewidth}
        \centering
        \resizebox{1\linewidth}{!}{%
        \begin{tabular}{lcc}
            \toprule
            & \cjc{G-VBSM (hrs)} & \cjc{CV-DD (hrs)} \\
            \midrule
            \cjc{IPC = 50}  & \cjc{187.5} & \cjc{137.5} \\
            \cjc{IPC = 100} & \cjc{375.0} & \cjc{190.3} \\
            \cjc{IPC = 150} & \cjc{562.5} & \cjc{243.1} \\
            \cjc{IPC = 200} & \cjc{750.0} & \cjc{295.9} \\
            \bottomrule
        \end{tabular}}
        \caption{\cjc{Total distillation time.}}
        \label{tab:total-distillation-time}
    \end{subtable}
    \vspace{-.1in}
\end{table*}

\begin{wraptable}{r}{0.5\columnwidth}
    \vspace{-.18in}
    \caption{Efficiency comparison of ensemble-based methods. Time per image per iteration is measured on an RTX-4090 with batch size 100 using identical committee models; N/A denotes non-scalable methods.}
    \vspace{-.05in}
    \label{tab:eff_compare}
    \centering
    \renewcommand{\arraystretch}{1}
    \resizebox{1\linewidth}{!}{  
        \begin{tabular}{ccccc}
            \toprule
            MTT & G-VBSM & EDC & \name{} \\
            \midrule 
             N/A & 4.32 ms & 4.99 ms & \textbf{1.91} ms \\ 
            \bottomrule
        \end{tabular}
    }
    \vspace{-.05in}
\end{wraptable}
\textbf{Efficiency Analysis.} We compare the efficiency of \name{} with ensemble methods on ImageNet-1K (Table~\ref{tab:eff_compare}). MTT incurs the highest cost and scales poorly. In contrast, \name{} is 2.41\,ms faster per iteration than G-VBSM, which introduces additional overhead from convolutional statistic alignment, similar to EDC, highlighting its computational advantage. We further evaluate the total prior-evaluation cost, including committee pretraining, distillation, and evaluation. As shown in Table~\ref{tab:prior-evaluation-time}, this stage requires 84.7 hours on ImageNet-1K, yielding 137.5 hours to distill a 50-IPC dataset, compared to 187.5 hours for G-VBSM (Table~\ref{tab:total-distillation-time}). Once completed, prior evaluation can be reused for subsequent runs, enabling faster synthesis.

\begin{figure*}[t]
    \centering
    \includegraphics[width=0.9\linewidth]{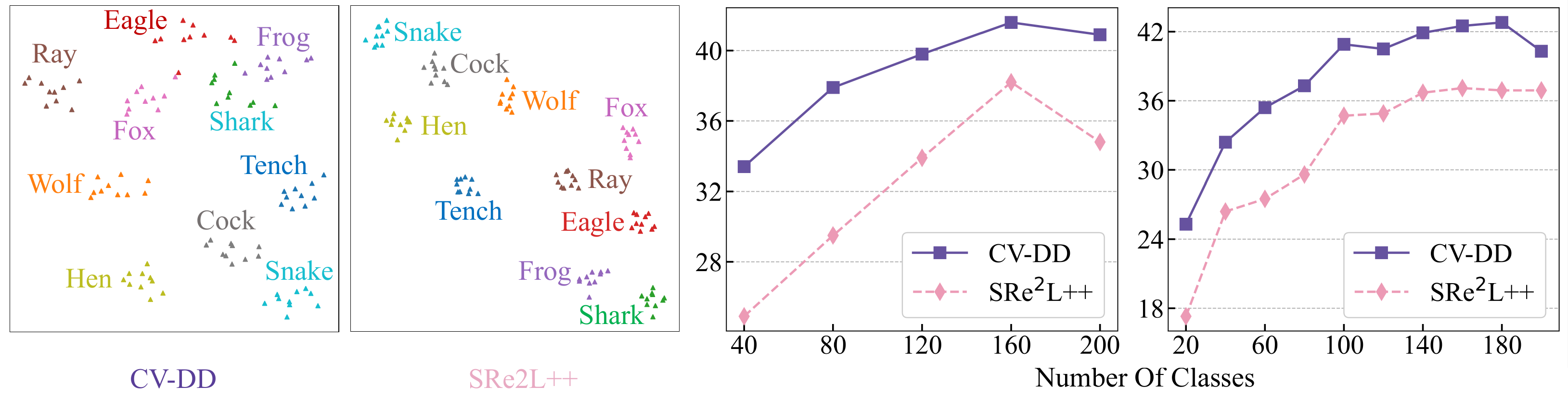}
    \vspace{-0.05in}
    \caption{\textbf{Left}: t-SNE visualization on ImageNet-1K (IPC=10) for ten selected classes. \textbf{Right}:  Continual learning results on Tiny-ImageNet (IPC=50) with 5 and 10 steps.}
    \label{fig:tsne_cl}
    \vspace{-1em}
\end{figure*}

\subsection{Application: Continual Learning}

Generalization in continual learning reflects distillation effectiveness. Under the class-incremental protocol of DM~\citep{DBLP:conf/wacv/ZhaoB23}, we evaluate \name{} in a continual learning setting. As shown in Fig.~\ref{fig:tsne_cl}, \name{} outperforms SRe$^2$L++, demonstrating robustness under distributional shifts and evolving tasks.

\subsection{Visualization}

\begin{figure}[t]
    \centering
    \includegraphics[width=0.75\linewidth]{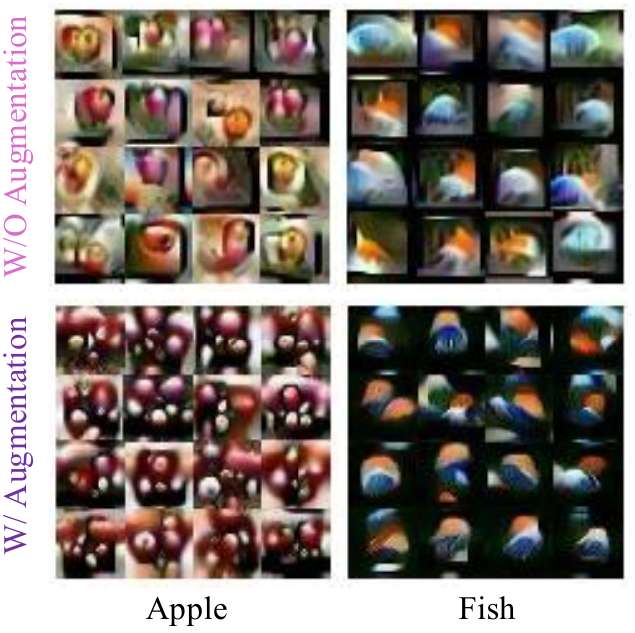}
    \vspace{-0.1in}
    \caption{Comparison of distilled data on CIFAR-100 generated by SRe$^2$L++ with and without data augmentation.}
    \label{fig:augmentation}
    \vspace{-.2in}
\end{figure}

\begin{figure}[t]
    \centering
    \includegraphics[width=0.88\linewidth]{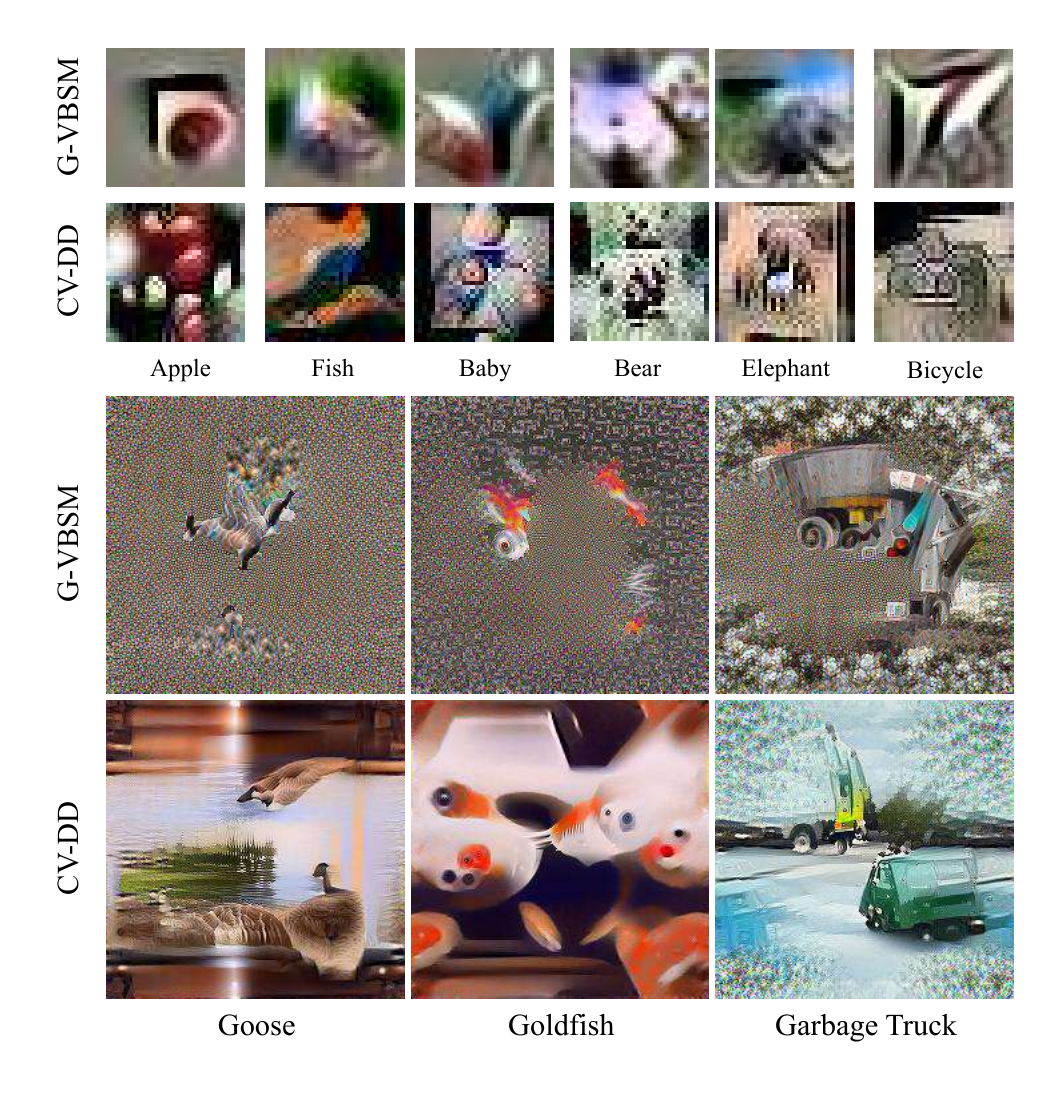}
    \caption{Top rows: CIFAR-100; Bottom rows: ImageNet-1K.}
    \label{fig:visualization-compare}
    \vspace{-.2in}
\end{figure}

\textbf{Effect of Data Augmentation.}
As shown in Fig.~\ref{fig:augmentation}, data augmentation enhances target saliency and information density, leading to improved quality of the distilled dataset.

\textbf{t-SNE Visualization.}
We extract ResNet-18 features for t-SNE~\citep{van2008visualizing} visualization (Fig.~\ref{fig:tsne_cl}). SRe$^2$L++ exhibits clustered features, indicating limited diversity, whereas \name{} yields more dispersed embeddings, reflecting greater variability and representational richness.

\textbf{Distilled Data Visualization.} Fig.~\ref{fig:visualization-compare} compares distilled data from G-VBSM and \name{} on CIFAR-100 and ImageNet-1K. G-VBSM shows lower overall information density, with ImageNet-1K samples often resembling noise. In contrast, \name{} captures more primary visual features, especially on CIFAR-100 with augmentation, yielding more informative synthetic samples.
\vspace{-0.05in}
\section{Conclusion}
\label{sec:conclusion}
We proposed \emph{Committee Voting} for dataset distillation (\name{}), a novel framework that synthesizes high-quality distilled datasets by leveraging prior performance guided voting strategy for image generation and batch-specific soft labeling for high-quality supervisions. Our approach first establishes a strong baseline that achieves state-of-the-art accuracy through recent advancements and carefully optimized framework design. By combining the distributions and predictions from a committee of models, our method captures rich data features, reduces model-specific biases, and enhances generalization. Complementing this, the generation of high-quality soft labels provides precise supervisory signals, effectively mitigating distribution shifts.  Building on these strengths, \name{} consistently improves performance across various configurations and datasets.

\section*{Impact Statement}
This work aims to improve the reliability and generalization of dataset distillation by leveraging committee voting and high-quality soft-label supervision. By reducing model-specific bias and mitigating distribution shift, our method enables more robust distilled datasets that can lower training cost, memory usage, and energy consumption, thereby facilitating broader access to efficient model training in resource-constrained settings. Potential risks include propagating or amplifying biases present in the underlying pretrained models used in the committee. We therefore encourage careful selection of committee members, transparent reporting of experimental settings, and thorough evaluation across diverse architectures and datasets when deploying the proposed approach.

\section*{Acknowledgements}
This work was supported by the MBZUAI-WIS Joint Program for AI Research.

\bibliography{main}
\bibliographystyle{icml2026}

\clearpage
\newpage
\appendix
\onecolumn

{	\Large \bf {Appendix}}

\section{Theoretical Analysis}
\subsection{Committee Diversity Enhances Data Diversity}

\begin{proof}[Proof of Theorem~\ref{thm:diversity-separation}]
\label{proof:committee_diversity}
We analyze the effect of committee diversity on intra-class separation via the dynamics of gradient-based updates to synthetic images.

Let \( \hat{u}_z^{(t)} \in \mathbb{R}^d \) denote the image for sample \( z \) at iteration \( t \), with update rule:
\[
\hat{u}_z^{(t+1)} = \hat{u}_z^{(t)} - \eta G_z^{(t)},
\]
where \( G_z^{(t)} = \sum_{i} w_i \nabla_{\hat{u}_z} \ell(\Phi_i(\hat{u}_z), \hat{v}_z) \) is the weighted ensemble gradient from the committee \( \{\Phi_i\} \). We assume all samples are \textbf{unit-normalized}, i.e., \( \|\hat{u}_z^{(t)}\| = 1 \).

Let \( d^{(t)} := \hat{u}_z^{(t)} - \hat{u}_{z'}^{(t)} \) denote the difference between two intra-class samples. Then,
\[
\begin{aligned}
d^{(t+1)} &= \hat{u}_z^{(t+1)} - \hat{u}_{z'}^{(t+1)} \\
&= \hat{u}_z^{(t)} - \eta G_z^{(t)} - \left( \hat{u}_{z'}^{(t)} - \eta G_{z'}^{(t)} \right) \\
&= d^{(t)} - \eta (G_z^{(t)} - G_{z'}^{(t)}).
\end{aligned}
\]

Taking the squared norm, we have:
\[
\begin{aligned}
\|d^{(t+1)}\|^2 &= \|d^{(t)} - \eta (G_z^{(t)} - G_{z'}^{(t)})\|^2 \\
&= \|d^{(t)}\|^2 + \eta^2 \|G_z^{(t)} - G_{z'}^{(t)}\|^2  \\ 
& - 2\eta \langle d^{(t)}, G_z^{(t)} - G_{z'}^{(t)} \rangle.
\end{aligned}
\]

Now taking expectation over intra-class pairs \( (z, z') \), we obtain:
\begin{align*}
    \mathbb{E}[\|d^{(t+1)}\|^2] &= \mathbb{E}[\|d^{(t)}\|^2] + \eta^2 \mathbb{E}[\|G_z^{(t)} - G_{z'}^{(t)}\|^2] \\
    &- 2\eta \mathbb{E}[\langle d^{(t)}, G_z^{(t)} - G_{z'}^{(t)} \rangle].
\end{align*}
We conservatively drop the final inner product term by assuming that the sample difference direction is uncorrelated with the gradient difference direction, and proceed using Assumption (ii):
\[
\mathbb{E}[\|d^{(t+1)}\|^2] \ge \mathbb{E}[\|d^{(t)}\|^2] + \eta^2 C_g K.
\]

Now, we relate this result to cosine distance. Under the unit norm assumption, we have:
\[
\begin{aligned}
\|\hat{u}_z^{(t)} - \hat{u}_{z'}^{(t)}\|^2 &= \|\hat{u}_z^{(t)}\|^2 + \|\hat{u}_{z'}^{(t)}\|^2 - 2\langle \hat{u}_z^{(t)}, \hat{u}_{z'}^{(t)} \rangle \\
&= 2(1 - \langle \hat{u}_z^{(t)}, \hat{u}_{z'}^{(t)} \rangle) \\
&= 2 \Delta_{\cos}^{(t)}.
\end{aligned}
\]

Thus, \( \Delta_{\cos}^{(t)} = \frac{1}{2} \|d^{(t)}\|^2 \), and we conclude:
\[
\begin{aligned}
\mathbb{E}[\Delta_{\cos}^{(t+1)}] &= \frac{1}{2} \mathbb{E}[\|d^{(t+1)}\|^2] \\
&\ge \frac{1}{2} \left( \mathbb{E}[\|d^{(t)}\|^2] + \eta^2 C_g K \right) \\
&= \mathbb{E}[\Delta_{\cos}^{(t)}] + \frac{1}{2} \eta^2 C_g K.
\end{aligned}
\]

This completes the proof: committee diversity \( K \) induces larger gradient differences, which in turn cause intra-class images to spread out in angular space, increasing \( \Delta_{\cos}^{(t)} \).
    
\end{proof}

\subsection{Prior Voting Enhances Distilled Data Quality}

\begin{proof}[Proof of Theorem~\ref{thrm:ppg is better}]
\label{proof:ppg is better}
Let \( J(\hat{u}) \) denote the generalization risk of a synthetic image \( \hat{u} \in \mathbb{R}^d \), and let \( \nabla J(\hat{u}) \) be its gradient. Suppose we have \( |S| \) models in the committee, where each model \( \Phi_i \in S \) is associated with a known prior performance score \( \alpha_i > 0 \).

Each model \( \Phi_i \) contributes a gradient:
\[
g_i := \nabla_{\hat{u}} \ell_{\Phi_i}(\hat{u}),
\]
where \( \ell_{\Phi_i}(\hat{u}) \) denotes the loss incurred by model \( \Phi_i \) on synthetic input \( \hat{u} \). We assume that the alignment between each model's gradient and the generalization gradient is given by:
\[
\left\langle \nabla J(\hat{u}), g_i \right\rangle = \lambda \cdot \alpha_i, \quad \text{for all } i \in \{1, \dots, |S|\},
\]
where \( \lambda > 0 \) is a constant.

We consider two aggregation strategies:

\textbf{(1) Equal voting:}
\[
G_{\text{equal}} := \frac{1}{|S|} \sum_{i=1}^{|S|} g_i.
\]

\textbf{(2) Prior-weighted voting using softmax over \( \alpha_i \):}
\[
w_i := \frac{\exp(\alpha_i)}{\sum_{j=1}^{|S|} \exp(\alpha_j)}, \quad
G_{\text{prior}} := \sum_{i=1}^{|S|} w_i g_i.
\]

We now compute the alignment between each aggregate direction and the generalization gradient.

\textbf{Step 1: Compute } \( \left\langle \nabla J(\hat{u}), G_{\text{equal}} \right\rangle \)

\begin{align*}
    \left\langle \nabla J(\hat{u}), G_{\text{equal}} \right\rangle 
&= \left\langle \nabla J(\hat{u}), \frac{1}{|S|} \sum_{i=1}^{|S|} g_i \right\rangle  \\
&= \frac{1}{|S|} \sum_{i=1}^{|S|} \left\langle \nabla J(\hat{u}), g_i \right\rangle  \\
&= \frac{1}{|S|} \sum_{i=1}^{|S|} \lambda \alpha_i \\
&= \lambda \cdot \bar{\alpha},
\end{align*}

where \( \bar{\alpha} := \frac{1}{|S|} \sum_{i=1}^{|S|} \alpha_i \) is the uniform average of prior performance.

\textbf{Step 2: Compute } \( \left\langle \nabla J(\hat{u}), G_{\text{prior}} \right\rangle \)

\begin{align*}
    \left\langle \nabla J(\hat{u}), G_{\text{prior}} \right\rangle 
& = \left\langle \nabla J(\hat{u}), \sum_{i=1}^{|S|} w_i g_i \right\rangle \\
& = \sum_{i=1}^{|S|} w_i \left\langle \nabla J(\hat{u}), g_i \right\rangle \\
& = \sum_{i=1}^{|S|} w_i \cdot \lambda \alpha_i \\
& = \lambda \cdot \sum_{i=1}^{|S|} w_i \alpha_i. 
\end{align*}

Let us denote the softmax-weighted average as:
\[
\mathbb{E}_w[\alpha] := \sum_{i=1}^{|S|} w_i \alpha_i,
\]
then we have:
\[
\left\langle \nabla J(\hat{u}), G_{\text{prior}} \right\rangle = \lambda \cdot \mathbb{E}_w[\alpha].
\]

\noindent\textbf{Step 3: Show } \( \mathbb{E}_w[\alpha] > \bar{\alpha} \)

\noindent Since \( \{w_i\} \) is a softmax distribution over \( \{\alpha_i\} \), we can write:
\[
w_i = \frac{\exp(\alpha_i)}{\sum_{j=1}^{|S|} \exp(\alpha_j)} \quad \text{for all } i.
\]
It assigns greater weight to higher values of \( \alpha_i \) in a strictly convex manner. Therefore, unless all \( \alpha_i \) are equal (which we exclude), the softmax-weighted average exceeds the arithmetic mean:
\[
\mathbb{E}_w[\alpha] > \bar{\alpha}.
\]

\noindent\textbf{Conclusion:}

Combining the results:
\[
\left\langle \nabla J(\hat{u}), G_{\text{prior}} \right\rangle 
= \lambda \cdot \mathbb{E}_w[\alpha] 
> \lambda \cdot \bar{\alpha} 
= \left\langle \nabla J(\hat{u}), G_{\text{equal}} \right\rangle,
\]
which shows that prior-weighted voting provides a descent direction that more closely aligns with the gradient of generalization risk, thereby confirming the theorem.
\end{proof}

\section{Additional Algorithmic Details}

\subsection{Optimization Details}
At each iteration \( t \), the optimization aims to update the synthetic image \( \hat{u}_z^{(t)} \) by minimizing an objective that jointly enforces prediction alignment and distributional consistency with respect to a fixed pretrained model \( \Phi \). The objective is given by:
\begin{align}
    \hat{u}_z^{(t+1)} \leftarrow \arg\min_{\hat{u}} \; \mathcal{L}_{\text{pred}}(\Phi(\hat{u}), \hat{v}_z) + \lambda \mathcal{L}_{\text{BN}}(\hat{u}),
\end{align}
where \( \mathcal{L}_{\text{pred}} \) ensures semantic alignment with the soft label \( \hat{v}_z \), and \( \mathcal{L}_{\text{BN}} \) regularizes the feature distribution of \( \hat{u} \) to match the population statistics of the original dataset. The coefficient \( \lambda \) balances the two objectives.

\noindent\textbf{Prediction Alignment Loss.}
The prediction loss is defined as cross-entropy between the model prediction and the soft label:
\begin{equation}
    \mathcal{L}_{\text{pred}}(\Phi(\hat{u}), \hat{v}) = - \sum_{c=1}^{C} \hat{v}_{c} \log \Phi(\hat{u})_c,
\end{equation}
where \( C \) is the number of classes and \( \Phi(\hat{u})_c \) denotes the predicted probability for class \( c \).

\noindent\textbf{BatchNorm Statistic Matching.}
To mitigate distributional shift between synthetic and real data, we introduce a regularization term that penalizes discrepancies in BatchNorm statistics:
\begin{align}
    \mathcal{L}_{\text{BN}}(\hat{u}) &= \sum_{l} \left\| \mu_l(\hat{u}) - \text{BN}^{\text{RM}}_l \right\|_2^2 
    + \sum_{l} \left\| \sigma^2_l(\hat{u}) - \text{BN}^{\text{RV}}_l \right\|_2^2,
\end{align}
where \( \mu_l(\hat{u}) \) and \( \sigma^2_l(\hat{u}) \) are the empirical mean and variance of the activations at layer \( l \) induced by the synthetic batch \( \hat{u} \), and \( \text{BN}^{\text{RM}}_l \), \( \text{BN}^{\text{RV}}_l \) denote the pretrained running mean and variance recorded on the original dataset. These statistics are frozen and not updated during synthesis. This formulation enforces that synthetic samples activate the network in a manner consistent with the original data distribution, facilitating better generalization under limited supervision.

\subsection{Detailed Voting Process}
\noindent\textbf{Voting in Image Optimization.}

\begin{algorithm}[H]
\caption{Prior Performance Guided Voting Strategy}
\label{alg:cv}
\begin{algorithmic}[1]
\Require Prior Performance $\alpha$, Committee Members $S$, Iteration Budgets $t$ 
\Ensure Synthetic Images $\hat{u}$
\State Initialize Synthetic Images $\hat{u}$
\State $S^{I_{2}^{1}}, S^{I_{2}^{2}} \gets \text{sample}(S, 2)$; \Comment{Randomly sample two committee members}
\State $\alpha^{I_{2}^{1}} \gets \alpha[S^{I_{2}^{1}}]$, $\alpha^{I_{2}^{2}} \gets \alpha[S^{I_{2}^{2}}]$; \Comment{Retrieve prior performance scores}
\For{$i = 1$ to $t$}
    \State Compute weighted loss: 
    \[
    \mathcal{L} \gets \sum_{j=1}^{2} \frac{\exp(\alpha^{I_{2}^{j}} / T)}{\sum_{k=1}^{2} \exp(\alpha^{I_{2}^{k}} / T)} \cdot \mathcal{L}_{S^{I_{2}^{j}}}(\hat{u})
    \]
    \State Perform backpropagation on $\mathcal{L}$
    \State Update synthetic images $\hat{u}$ via gradient descent
\EndFor
\State Return optimized images $\hat{u}$
\end{algorithmic}
\end{algorithm}

\noindent\textbf{Voting in Soft Label Generation.}

\begin{algorithm}[H]
\caption{Prior-Guided Soft Label Aggregation}
\label{alg:soft_label_full}
\begin{algorithmic}[1]
\Require Committee \( \mathcal{S} = \{\Phi_1, \ldots, \Phi_{|\mathcal{S}|} \} \), prior performance \( \{ \alpha_i \}_{i=1}^{|\mathcal{S}|} \), temperature \( T \), synthetic sample \( \hat{u} \in \mathbb{R}^{C \times H \times W} \)
\Ensure Soft label \( \hat{v}  \)

\State Compute prior-based weights via temperature-scaled SoftMax:
\[
w_i \gets \frac{\exp(\alpha_i / T)}{\sum_{j=1}^{|\mathcal{S}|} \exp(\alpha_j / T)} \quad \text{for } i = 1, \ldots, |\mathcal{S}|
\]
\State Compute class probabilities from each model:
\[
p_i \gets \Phi_i(\hat{u})
\]
\State Aggregate weighted soft label:
\[
\hat{v} \gets \sum_{i=1}^{|\mathcal{S}|} w_i \cdot p_i
\]
\State Return \( \hat{v} \)
\end{algorithmic}
\end{algorithm}

\section{\cjc{Additional Experiments}}

\subsection{\cjc{Comparison between SRe$^2$L++ and SRe$^2$L}}

\cjc{In this section, we present the quantitative improvements of SRe$^{2}$L++ over SRe$^{2}$L on both large- and small-scale datasets, as summarized in Table~\ref{tab:sre2l_improvement}. We observe that after incorporating the existing techniques described in Section~\ref{sec:strong_baseline} and our proposed BSSL, SRe$^{2}$L++ achieves substantial performance gains across all settings.}

\begin{table}[htbp]
\centering
\caption{\cjc{Comparison of SRe$^2$L++ with original SRe$^2$L across datasets with ResNet18. Best in each row is \textbf{bold}.}}
\label{tab:sre2l_improvement}
\resizebox{0.45\linewidth}{!}{  
\begin{tabular}{llcc}
\toprule
Dataset &  IPC (Ratio)  & SRe$^{2}$L  &  SRe$^{2}$L++  \\
\midrule
\multirow{2}{*}{CIFAR-100} &
10 (2.0\%)  & 23.5  & \textbf{56.7}   \\
& 50 (10.0\%) & 51.4  &  \textbf{66.6}\\
\midrule
\multirow{1}{*}{Tiny-ImageNet} &
 50 (10.0\%) & 41.1  &  \textbf{46.5} \\
\midrule
\multirow{2}{*}{ImageNet-1k} &
10 (0.8\%)  &  21.3 &\textbf{43.1} \\
& 50 (3.9\%)  & 46.8  & \textbf{57.6}  \\
\bottomrule
\end{tabular}
}

\label{tab:cvdd_main}
\end{table}

\subsection{\cjc{Performance Gain from Different Components}}

\cjc{In this part, we explicitly quantify the performance gains contributed by each component of the enhanced SRe$^2$L++ baseline. As shown in Table~\ref{tab:ablation-components}, we progressively add individual techniques and measure their impact. Among all components, the improvement brought by using a small batch size is the most significant.}

\begin{table}[htbp]
\centering
\caption{\cjc{Ablation study of components contributing to SRe$^2$L++ performance.}}
\label{tab:ablation-components}
\resizebox{0.8\textwidth}{!}{
\begin{tabular}{ccccc|c}
\toprule
\textbf{Data Augmentation} & 
\textbf{Real Data Initialization} & 
\textbf{Small Batch Size} & 
\textbf{Smoothed LR} & 
\textbf{BSSL} & 
\textbf{Performance} \\
\midrule
--            & --            & --            & --            & --            & \cjc{23.3} \\
\checkmark    & --            & --            & --            & --            & \cjc{30.8} \\
\checkmark    & \checkmark    & --            & --            & --            & \cjc{31.3} \\
\checkmark    & \checkmark    & \checkmark    & --            & --            & \cjc{37.6} \\
\checkmark    & \checkmark    & \checkmark    & \checkmark    & --            & \cjc{38.5} \\
\checkmark    & \checkmark    & \checkmark    & \checkmark    & \checkmark    & \cjc{43.1} \\
\bottomrule
\end{tabular}
}
\end{table}

\subsection{\cjc{Impact of Online Reweighting}}

\textbf{}\cjc{In this part, we introduce an additional baseline that incorporates online reweighting within CV-DD, and compare it against CV-DD with fixed priors. Specifically, under the online reweighting scheme, the two experts start with equal weights of 0.5, and the weights are gradually updated toward their true prior values, reaching the final prior only in the last iteration. We then evaluate the quality of the distilled dataset produced under this strategy.}
\cjc{As shown in Table~\ref{tab:online-vs-fixed} below, the online reweighting approach yields lower performance compared to CV-DD (fixed priors). This degradation occurs because initializing both experts at 0.5 ignores the advantage of the stronger teacher during the early stages of optimization, leading to a suboptimal trajectory and ultimately a lower-quality distilled dataset.}

\begin{table}[htbp]
\caption{\cjc{Comparison between online reweighting and fixed-prior CV-DD.}}
\label{tab:online-vs-fixed}
\centering
\resizebox{0.6\linewidth}{!}{
\begin{tabular}{lcc}
\toprule
& \textbf{\cjc{CV-DD (Online Reweighting)}} & \textbf{\cjc{CV-DD (Fixed Prior)}} \\
\midrule
\textbf{\cjc{Performance}} & \cjc{48.3} & \cjc{49.5} \\
\bottomrule
\end{tabular}
}
\end{table}

\subsection{\cjc{Impact of Gradient Aggregation}}

\textbf{}\cjc{To examine whether CV-DD’s can be benefited from partial gradient aggregation, we evaluate a concrete variant of CV-DD where we compute full prediction alignment (cross-entropy) for all experts, but compute the distribution-alignment gradient for only a single expert. This setting reduces the per-iteration computation while retaining part of the supervisory signal used in CV-DD.} \cjc{The per-iteration speedup and the corresponding post-evaluation performance under this partial-gradient setting are reported in Table~\ref{tab:partial-vs-full}. We observe that, although this approach substantially lowers the computational cost, it also leads to a noticeable degradation in the quality of the distilled dataset. Since CV-DD is already considerably more efficient than existing ensemble-based methods, the additional benefits of adopting partial gradient aggregation remain limited.}

\begin{table}[htbp]
\caption{Comparison between partial and full gradient aggregation.}
\label{tab:partial-vs-full}
\centering
\resizebox{0.5\linewidth}{!}{
\begin{tabular}{l|cc}
\toprule
 & \cjc{\textbf{Per-iteration Cost}} & \cjc{\textbf{Performance}} \\
\midrule
\cjc{\textbf{Partial Gradients}} & \cjc{0.96 ms} & \cjc{47.1} \\
\cjc{\textbf{Full Gradients}}    & \cjc{1.91 ms} & \cjc{\textbf{49.5}} \\
\bottomrule
\end{tabular}
}
\end{table}

\section{\cjc{Incorporating Committee Voting to RDED}}

\cjc{To examine whether the committee-voting idea generalizes to other non-optimization based methods, we further incorporate \textbf{CV-DD} into \textbf{RDED}, where the original RDED selects patches from the dataset using a single model. Its selection function is defined as:}
\begin{equation}
\xi_{i,\star} = \arg\max_{\xi_{i,k} \sim p(\xi_{i,k} \mid \hat{x}_i)}
\left[-\ell\big(\phi_{\theta_T}(\xi_{i,k}), \phi_h(\xi_{i,k})\big)\right].
\end{equation}

\cjc{We extend this formulation by employing multiple models to select patches and weighting their losses according to each model’s prior performance:}
\begin{equation}
\xi_{i,\star} = 
\arg\max_{\xi_{i,k} \sim p(\xi_{i,k} \mid \hat{x}_i)}
\left[
-\sum_{m=1}^{n} 
w_m \, \ell\big(\phi_{\theta_m}(\xi_{i,k}), \phi_h(\xi_{i,k})\big)
\right],
\end{equation}
\cjc{where} 
\begin{equation}
w_m = \frac{\exp(\alpha_m / T)}{\sum_{j=1}^{n} \exp(\alpha_j / T)}.
\end{equation}

\section{Experimental Setup}

We provide detailed information regarding the hardware, software environment, and reproducibility measures to facilitate faithful replication of our experiments.

\begin{itemize}
    \item \textbf{Hardware.} All experiments were conducted on a dedicated Linux server equipped with 6 NVIDIA RTX 4090 GPUs (each with 24GB memory), 256GB of RAM, and an AMD Ryzen Threadripper PRO 5995WX CPU. The operating system was Ubuntu 22.04.5 LTS. This high-performance computing environment ensured stable and efficient execution of all training and evaluation procedures.

    \item \textbf{Software.} The experiments were implemented using PyTorch version 2.0.1, with CUDA 11.8 for GPU acceleration and Python 3.10 as the primary programming language. Data augmentation pipelines were constructed using the \texttt{torchvision} library, and all experiment tracking and logging were managed via the \texttt{wandb} platform. All dependencies and library versions are specified in the provided codebase to ensure compatibility and reproducibility.

    \item \textbf{Reproducibility.} To ensure deterministic behavior across runs, we explicitly set random seeds. All hyperparameter settings used in our experiments are documented in Appendix~\ref{sec:hyper}, and full implementation details, including scripts for data preprocessing, training, and evaluation, are provided in the supplementary material.

    \item \textbf{Experiment Execution.} To ensure the robustness and reliability of our findings, all reported results are averaged over three independent runs. Each run follows the exact same pipeline, including data loading, model initialization, training, and evaluation, under controlled randomization settings. This repeated execution mitigates the impact of stochasticity inherent in deep learning training processes and provides a more stable estimate of performance.
    
\end{itemize}

\section{Prior Performance}
\textbf{Prior Performance Computation.} This section provides a detailed account of the prior performance of each model across a range of benchmark datasets. \cjc{We generate a fixed set of 10 synthetic images per class (IPC = 10) for ImageNet-1K and 50 per class (IPC = 50) for other datasets. Each model trained on the distilled images is then evaluated on its corresponding validation set, and the Top-1 classification accuracy is reported.} As shown in Table~\ref{tab:prior-performance}, the resulting performance scores serve as a proxy for each model’s generalization ability and form the basis for subsequent ensemble weighting and soft label refinement. Importantly, we fix the IPC across all models to ensure that their prior performance scores are within a comparable range. This is particularly critical because the voting temperature parameter $T$ is shared across models (with $T = 5$ by default), and having performance values on a similar scale avoids skewing the voting distribution due to imbalanced score magnitudes.

\begin{table*}[htbp]
\centering
\caption{Prior Performance for different models across different datasets.}
\label{tab:prior-performance}
\resizebox{0.8\linewidth}{!}{
\begin{tabular}{lcccccc}
\toprule
Model & CIFAR-10 & CIFAR-100 & Tiny-ImageNet & ImageNette & ImageWoof & ImageNet-1K \\
\midrule
ResNet18       & 75.41 & 66.47 & 56.34 & 81.22 & 63.8 & \cjc{42.3} \\
ResNet50       & 70.59 & 66.36 & 56.02 & 73.01 & 50.9& \cjc{31.4} \\
DenseNet121    & 75.46 & 65.62 & 56.81 & 77.83 & 63.3  & \cjc{34.6} \\
ShuffleNetV2   & 52.63 & 57.22 & 55.41 & 71.47 &43.1  & \cjc{39.3} \\
MobileNetV2    & 71.37 & 65.03 & 56.43 & 72.12 & 53.8 & \cjc{33.2} \\
\bottomrule
\end{tabular}}
\end{table*}

\section{Hyperparameter Configuration}
\label{sec:hyper}
The overall synthetic data generation process adheres to a unified hyperparameter configuration across all experimental settings, as summarized in Table~\ref{tab:recover}. This consistency facilitates fair comparisons and controlled ablation studies. Deviations from this configuration are restricted to two specific stages: (i) the post-evaluation phase, and (ii) the supervised pre-training of committee members. In these stages, hyperparameters are selectively adjusted to accommodate architectural differences in model capacity and the resolution characteristics of the target dataset.

Specifically, during the pre-training phase, we train five distinct architectures on each dataset to construct the committee: ResNet18, ResNet50, ShuffleNetV2, MobileNetV2, and DenseNet121. These models are chosen to span a range of parameter complexities and inductive biases, ensuring both architectural diversity and representational complementarity. Hyperparameters in this phase, such as learning rate schedules and batch sizes, are tuned with respect to each model's convergence behavior and computational profile.

\begin{table}[htbp]
\centering
\caption{Hyperparameters for generating synthetic data across all five datasets.} 
\label{tab:recover}
\begin{tabular}{cc} 
\toprule
\multicolumn{2}{c}{\textbf{Hyperparameters for Synthetic Data Generation}} \\
\midrule
Optimizer & Adam\\
Learning rate & 0.25 \\
beta & 0.5, 0.9 \\
epsilon & 1e-8 \\
Batch Size & 100 or 10 (if $C < 100$)\\
Iterations & 2,000 \\
Scheduler & Cosine Annealing \\
Augmentation & RandomResizedCrop, Horizontal Flip \\
\bottomrule
\end{tabular}
\end{table}

\subsection{CIFAR-10}
This subsection delineates the complete set of hyperparameter configurations employed in our experiments on CIFAR-10, with the goal of facilitating rigorous reproducibility and enabling future benchmarking efforts under consistent conditions.

\noindent \textbf{Pre-training of Committee Models.} The detailed training configurations for all backbone models utilized in the committee are presented in Table~\ref{tab:squeeze_cifar10}. These models are trained on the full CIFAR-10 dataset using standard supervised learning protocols. The selection of hyperparameters, including learning rates, batch sizes, weight decay, and optimization schedules, has been empirically validated to ensure stable convergence and competitive generalization performance across architectures. The consistency of these training settings is critical for maintaining comparability across committee members and ensuring the integrity of the prior-performance estimation process.

\begin{table}[htbp]
\centering
\caption{Hyperparameters for CIFAR-10 Pre-trained Models.} 
\label{tab:squeeze_cifar10}
\begin{tabular}{cc} 
\toprule
\multicolumn{2}{c}{\textbf{Hyperparameters for Model Pre-training}} \\
\midrule
Optimizer & SGD\\
Learning rate & 0.01 \\
Weight Decay & 1e-4  \\
Batch Size & 64\\
Epoch & 10 \\
Scheduler & Cosine Annealing \\
Augmentation & RandomResizedCrop, Horizontal Flip \\
Loss Function & Cross-Entropy \\
\bottomrule
\end{tabular}
\end{table}

\noindent \textbf{Post-Evaluation Phase.} The hyperparameter settings adopted during the post-evaluation stage on the distilled CIFAR-10 dataset are comprehensively summarized in Table~\ref{tab:validate_cifar10}. These configurations govern the training of evaluation models on the synthetic data and are designed to reflect standard supervised learning protocols, thereby enabling fair comparisons across distilled datasets. Parameters such as learning rate schedules, number of training epochs, regularization strength, and batch size are carefully selected to ensure stable optimization dynamics while preserving sensitivity to differences in distilled data quality.

\begin{table}[htbp]
\centering
\caption{Hyperparameters for post-evaluation task on ResNet18, ResNet50 and ResNet101 for CIFAR-10.}
\label{tab:validate_cifar10}
\begin{tabular}{p{3.5cm} p{4.5cm}}
\toprule
\multicolumn{2}{c}{\textbf{Hyperparameters for Post-Eval on R18, R50 and R101}} \\
\midrule
Optimizer      & AdamW                         \\ 
Learning Rate  & 0.001 (ResNet18, ResNet50) or \newline 0.0005 (ResNet101)                        \\ 
Loss Function  & KL-Divergence                \\ 
Batch Size     & 16 or 10 (if $|S| \leq 16$)                           \\ 
Epochs         & 300                          \\
Scheduler      & Cosine Annealing with $\eta=1$ \\
Augmentation   & RandomResizedCrop, \newline Horizontal Flip, CutMix \\
\bottomrule
\end{tabular}
\end{table}

\subsection{CIFAR-100}

This subsection presents the complete hyperparameter configurations utilized in our CIFAR-100 experiments, offering a transparent and reproducible foundation for future research endeavors.

\noindent \textbf{Pre-training of Committee Models.} The hyperparameters employed for training the committee models on the original CIFAR-100 dataset are detailed in Table~\ref{tab:squeeze_cifar100}. Each model is trained under a consistent supervised learning framework, with dataset-specific adjustments to accommodate the increased complexity and fine-grained nature of CIFAR-100. The configuration includes optimization parameters such as initial learning rate, batch size, learning rate decay schedule, and regularization strength. These settings are empirically tuned to promote convergence stability and to ensure that the resulting models serve as reliable sources for prior-performance estimation within the committee-based distillation framework.

\begin{table}[t]
\centering
\caption{Hyperparameters for CIFAR-100 Pre-trained Models.} 
\label{tab:squeeze_cifar100}
\begin{tabular}{cc} 
\toprule
\multicolumn{2}{c}{\textbf{Hyperparameters for Model Pre-training}} \\
\midrule
Optimizer & SGD\\
Learning rate & 0.1 \\
Weight Decay & 1e-4  \\
Batch Size & 512\\
Epoch & 200 \\
Scheduler & Cosine Annealing \\
Augmentation & RandomResizedCrop, Horizontal Flip \\
Loss Function & Cross-Entropy \\
\bottomrule
\end{tabular}
\end{table}

\noindent \textbf{Post-Evaluation Phase.} The hyperparameter settings utilized during the post-evaluation stage on the distilled CIFAR-100 dataset are comprehensively summarized in Table~\ref{tab:validate_cifar100}. These configurations govern the training of evaluation models exclusively on the synthetic data and are selected to ensure a fair and controlled assessment of the generalization capability imparted by the distilled representations. Parameters such as learning rate, batch size, number of training epochs, and weight decay are carefully chosen to balance convergence speed and evaluation sensitivity, thereby providing a reliable basis for comparative analysis across distillation methods.

\begin{table}[t]
\centering
\caption{Hyperparameters for post-evaluation task on ResNet18, ResNet50 and ResNet101 for CIFAR-100.}
\label{tab:validate_cifar100}
\begin{tabular}{p{3.5cm} p{4.5cm}}
\toprule
\multicolumn{2}{c}{\textbf{Hyperparameters for Post-Eval on R18, R50 and R101}} \\
\midrule
Optimizer      & AdamW                         \\ 
Learning Rate  & 0.001 (ResNet18, ResNet50) or \newline  0.0005 (ResNet101)                \\ 
Loss Function  & KL-Divergence                \\ 
Batch Size     & 16                          \\ 
Epochs         & 300                        \\
Scheduler      & Cosine Annealing with  \newline $\eta=1$ (ResNet18, IPC=1, 50) or \newline $\eta=1$ (ResNet50, IPC=1, 50) or \newline $\eta=1$ (ResNet101) or \newline $\eta=2$ (ResNet18, IPC=10) or \newline $\eta=2$ (ResNet50, IPC=10) \\
Augmentation   & RandomResizedCrop, \newline Horizontal Flip, CutMix \\
\bottomrule
\end{tabular}
\end{table}

\subsection{Tiny-ImageNet}

\noindent \textbf{Pre-training of Committee Models.} Table~\ref{tab:squeeze_tiny} provides a comprehensive summary of the hyperparameter configurations adopted for training the backbone models on the original Tiny-ImageNet dataset. These models serve as committee members in the distillation framework, and their training is conducted under standardized supervised learning protocols. The selected hyperparameters, including optimizer choice, initial learning rate, batch size, weight decay, and learning rate scheduling, are tailored to accommodate the higher input resolution and increased class cardinality of Tiny-ImageNet, thereby ensuring robust convergence and high-quality feature representations for subsequent distillation.

\begin{table}[t]
\centering
\caption{Hyperparameters for Training Tiny-ImageNet Pre-trained Models.} 
\label{tab:squeeze_tiny}
\begin{tabular}{cc} 
\toprule
\multicolumn{2}{c}{\textbf{Hyperparameters for Model Pre-training}} \\
\midrule
Optimizer & SGD\\
Learning rate & 0.1 \\
Momentum & 0.9  \\
Batch Size & 128\\
Epoch & 50 \\
Scheduler & Cosine Annealing \\
Augmentation & RandomResizedCrop, Horizontal Flip \\
Loss Function & Cross-Entropy \\
\bottomrule
\end{tabular}
\end{table}

\noindent \textbf{Post-Evaluation Phase.} The hyperparameter configurations employed during the post-evaluation phase on the distilled Tiny-ImageNet dataset are detailed in Table~\ref{tab:validate_tiny}. These settings govern the training of evaluation models exclusively on the synthesized data and are selected to ensure a rigorous and standardized assessment of the distilled set's efficacy. Given the higher resolution and increased class diversity of Tiny-ImageNet, the configurations including learning rate schedule, regularization strategy, and optimization budget are carefully adapted to preserve stability during training while maintaining sufficient sensitivity to distinguish performance across distillation methods.

\begin{table}[t]
\centering
\caption{Hyperparameters for post-evaluation task on ResNet18, ResNet50 and ResNet101 for Tiny-ImageNet.}
\label{tab:validate_tiny}
\begin{tabular}{p{3.5cm} p{4.5cm}}
\toprule
\multicolumn{2}{c}{\textbf{Hyperparameters for Post-Eval on R18, R50 and R101}} \\
\midrule
Optimizer      & Adamw                         \\ 
Learning Rate  & 0.001 (ResNet18) or\newline 0.001 (ResNet50) or\newline 0.0005 (ResNet101)         \\ 
Loss Function  & KL-Divergence                \\ 
Batch Size     & 16                             \\ 
Epoch         & 300                          \\
Scheduler      & Cosine Annealing with \newline $\eta=1$ (ResNet18 IPC=50) or \newline $\eta=1$ (ResNet50) or\newline $\eta=2$  (ResNet18 IPC=1, 10) or\newline $\eta=2$ (ResNet101) \\
Augmentation   & RandomResizedCrop, \newline Horizontal Flip, CutMix \\
\bottomrule
\end{tabular}
\end{table}

\subsection{Subsets of ImageNet-1K}

\textbf{Pre-training of Committee Models.} The full set of training configurations used to pre-train committee models on the ImageNet-1K subsets is summarized in Table~\ref{tab:squeeze_imagenet_nette}. These configurations are tailored to accommodate the high-resolution inputs and the fine-grained semantic variability inherent to ImageNet class data. Optimization hyperparameters including learning rate initialization, batch size, weight decay, and scheduler policies are selected to ensure stable convergence across model architectures and to produce reliable feature extractors for use in the distillation process.

\begin{table}[t]
\centering
\caption{Hyperparameters for Training subsets of ImageNet-1k Pre-trained Models.} 
\label{tab:squeeze_imagenet_nette}
\begin{tabular}{cc} 
\toprule
\multicolumn{2}{c}{\textbf{Hyperparameters for Model Pre-training}} \\
\midrule
Optimizer & SGD\\
Learning rate & 0.01 \\
Momentum & 0.9 \\
weight decay & 1e-4 \\
Batch Size & 64\\
Epoch & 300 \\
Scheduler & Cosine Annealing \\
Augmentation & RandomResizedCrop, Horizontal Flip \\
Loss Function & Cross-Entropy \\
\bottomrule
\end{tabular}
\end{table}

\noindent\textbf{Post-Evaluation Phase.} The hyperparameter configurations employed during the post-evaluation phase on the distilled subsets of the ImageNet-1K dataset are presented in Table~\ref{tab:validate_imagenet_nette}. These settings define the training protocols for evaluating models exclusively on the synthesized data and are carefully selected to reflect standardized supervised learning procedures. Particular attention is given to the choice of optimization parameters, including learning rate schedules, regularization strategies, and training durations, in order to ensure a fair and sensitive evaluation of the generalization capacity induced by the distilled representations.

\begin{table}[t]
\centering
\caption{Hyperparameters for post-evaluation task on ResNet18, ResNet50 and ResNet101 for subsets of ImageNet-1k.}
\label{tab:validate_imagenet_nette}
\begin{tabular}{p{3.5cm} p{4.5cm}}
\toprule
\multicolumn{2}{c}{\textbf{Hyperparameters for post-eval on R18, R50 and R101}} \\
\midrule
Optimizer      & Adamw                         \\ 
Learning Rate  & 5e-4 (ResNet18) or \newline 5e-4 (ResNet50, IPC=1, 50) or \newline 5e-4 (ResNet101, IPC=1, 10) or \newline 0.001 (ResNet50, IPC=10)  or \newline 0.001 (ResNet101, IPC=50)                   \\ 
Loss Function  & KL-Divergence                \\ 
Batch Size     & 10                          \\ 
Epochs         & 300                          \\
Scheduler      & Cosine Annealing with \newline $\eta=1$ (ResNet18, IPC=10) or \newline $\eta=1$ (ResNet50, IPC=10, 50) or \newline $\eta=2$ (ResNet18, IPC=1, 50) or \newline $\eta=2$ (ResNet50, IPC=1)  or \newline $\eta=2$ (ResNet101)  \\
Augmentation   & RandomResizedCrop, \newline Horizontal Flip, CutMix \\
\bottomrule
\end{tabular}
\end{table}

\subsection{ImageNet-1K}
This subsection offers a comprehensive account of the hyperparameter configurations adopted in the ImageNet-1K experiments, with the objective of facilitating reproducibility and enabling consistent comparisons in future research. As the generation of distilled data relies on the use of publicly available ImageNet-1K models provided by the official PyTorch repository, no modifications are made to the pretraining phase. Consequently, only the hyperparameters associated with the post-evaluation stage are reported, as summarized in Table~\ref{tab:validate_imagenet1k}.

\begin{table}[t]
\centering
\caption{Hyperparameters for post-evaluation task on ResNet18, ResNet50 and ResNet101 for ImageNet-1K.}
\label{tab:validate_imagenet1k}
\begin{tabular}{p{3.5cm} p{4.5cm}}
\toprule
\multicolumn{2}{c}{\textbf{Hyperparameters for Post-Eval on R18, R50 and R101}} \\
\midrule
Optimizer      & Adamw                         \\ 
Learning Rate  & 0.0005(ResNet101, IPC=50) or 0.001                    \\ 
Loss Function  & KL-Divergence                \\ 
Batch Size     & 16                           \\ 
Epochs         & 300                          \\
Scheduler      & Cosine Annealing with \newline $\eta=1$ (ResNet18, IPC=50) or \newline $\eta=1$ (ResNet50, IPC=50) or \newline $\eta=2$ (ResNet18, IPC=1, 10) or \newline $\eta=2$ (ResNet50, IPC=1, 10) or  \newline $\eta=2$ (ResNet101)\\
Augmentation   & RandomResizedCrop, \newline Horizontal Flip, CutMix \\
\bottomrule
\end{tabular}
\end{table}

\subsection{Cross-Architecture Generalization}
This subsection provides a detailed exposition of the hyperparameter configurations employed in the cross-architecture generalization experiments, with the aim of promoting reproducibility and facilitating robust comparisons in future studies. To accommodate variations in model capacity, the experimental protocol differentiates between architectures based on their parameter scales. Specifically, architectures with relatively large parameter counts, such as DenseNet201, are assigned the same hyperparameter settings as those used for ResNet101. Conversely, lightweight models adopt the configurations associated with ResNet18 and ResNet50. The complete set of hyperparameters utilized across these architectural variants is documented in Table~\ref{tab:validate_imagenet1k}.

\newpage

\section{Distilled Data Visualization}
\begin{figure*}[!t]
    \centering
    \includegraphics[width=0.9\textwidth]{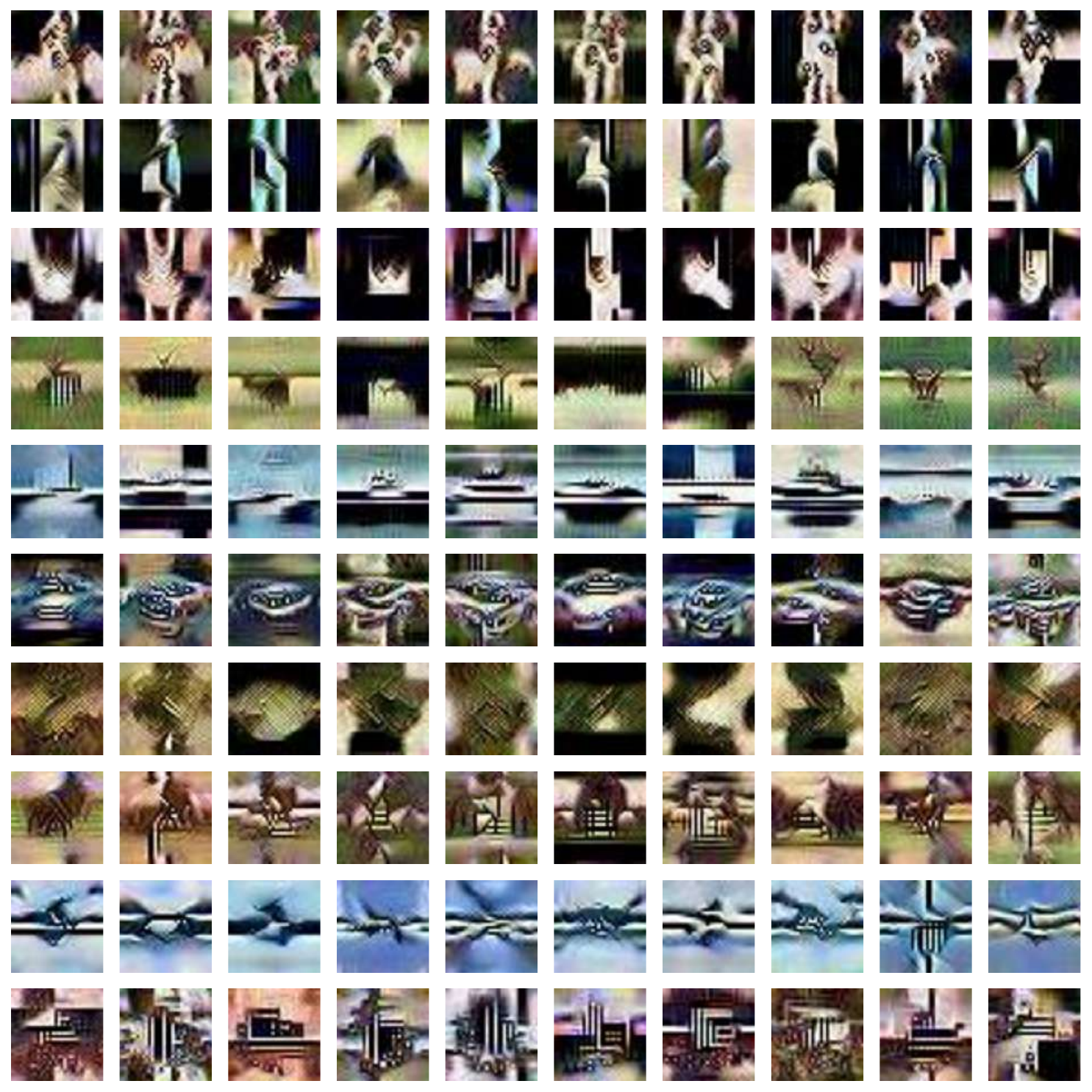}
    \caption{Visualization of synthetic data on CIFAR-10 generated by CV-DD.}
    \label{fig:cifar10_vis}
\end{figure*}

\begin{figure*}[!t]
    \centering
    \includegraphics[width=0.9\textwidth]{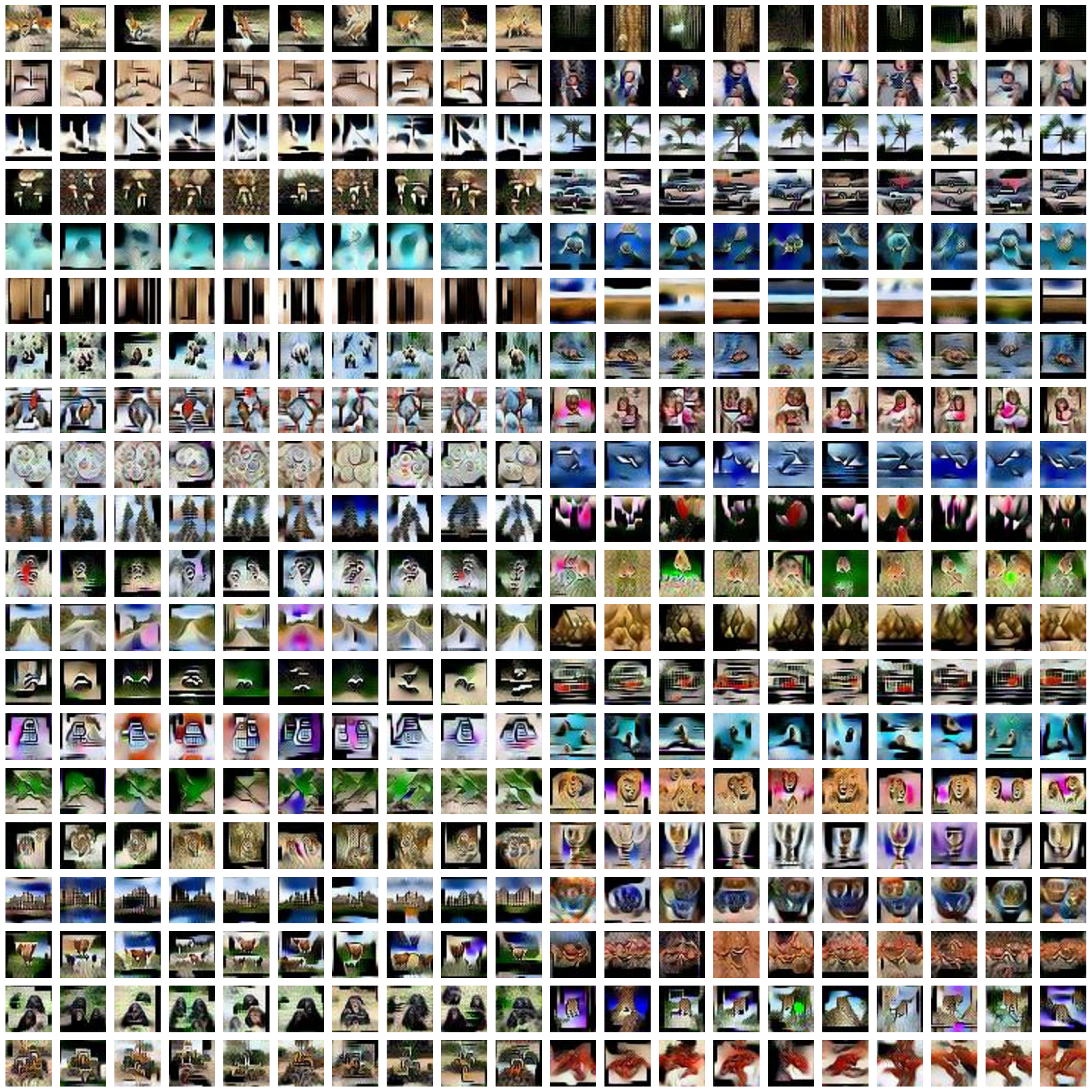}
    \caption{Visualization of synthetic data on CIFAR-100 generated by CV-DD.}
    \label{fig:cifar100_vis}
\end{figure*}

\begin{figure*}[!t]
    \centering
    \includegraphics[width=0.9\textwidth]{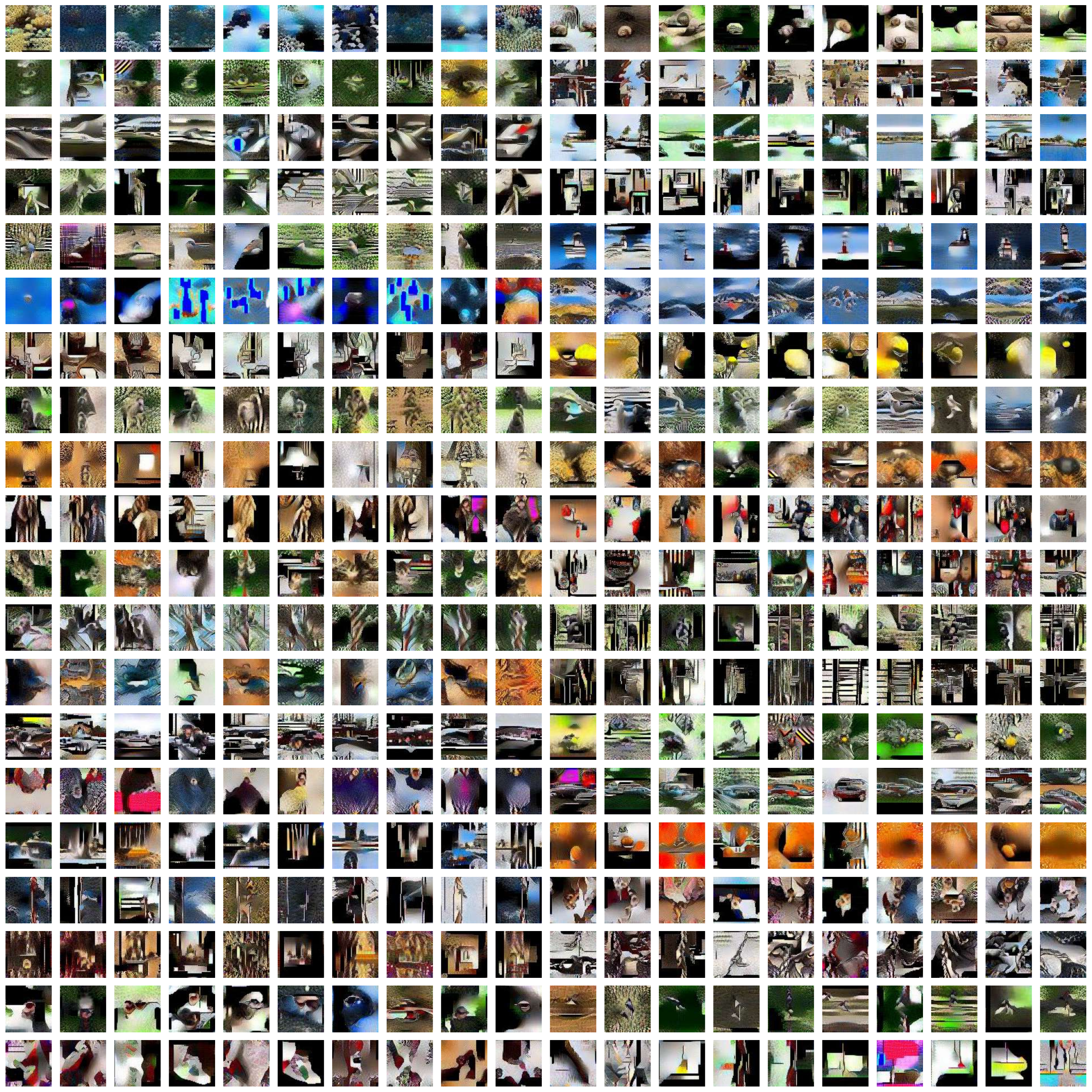}

    \caption{Visualization of synthetic data on Tiny-ImageNet generated by CV-DD.}
    \label{fig:tiny_vis}

\end{figure*}

\begin{figure*}[!t]
    \centering
    \includegraphics[width=0.9\textwidth]{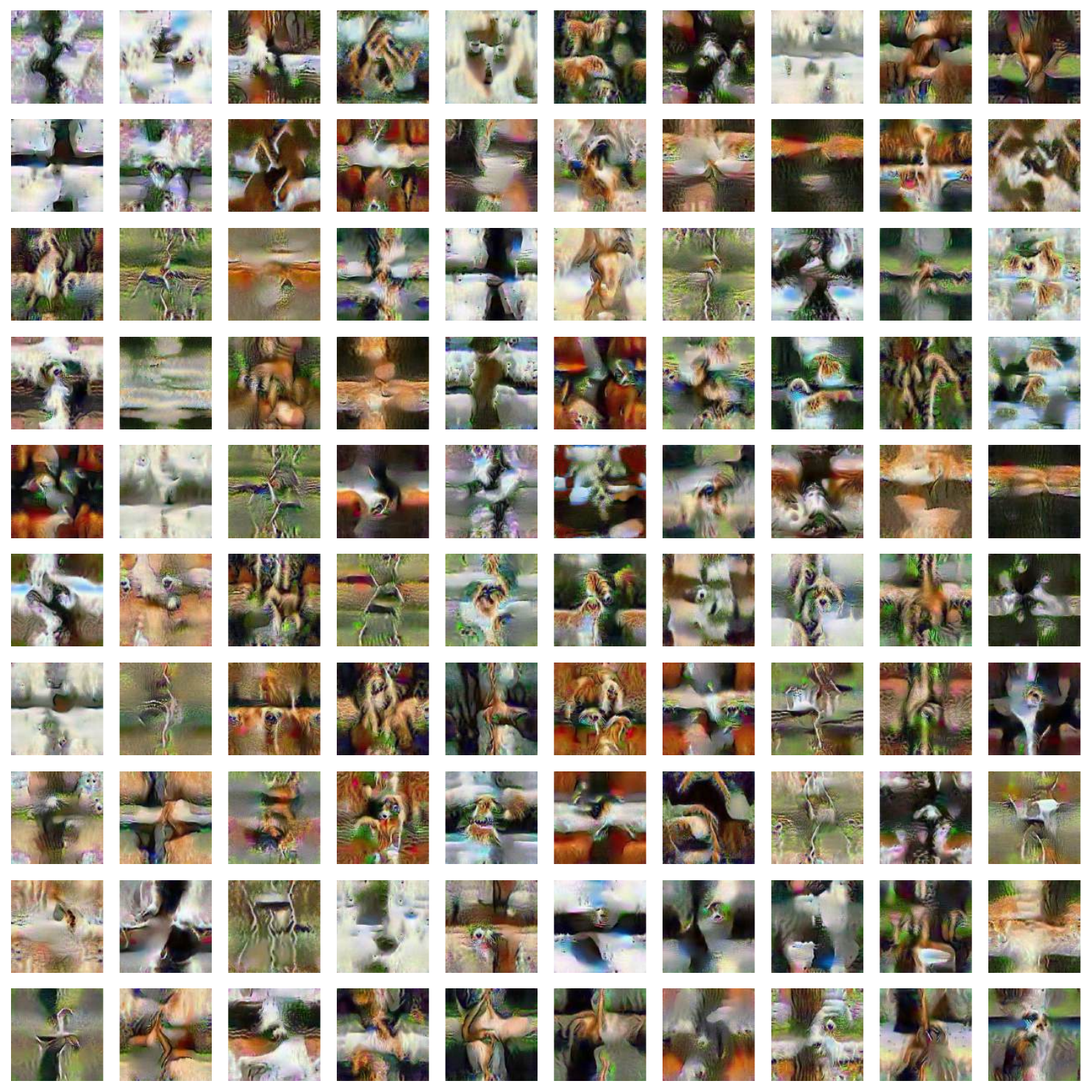}

    \caption{Visualization of synthetic data on ImageWoof generated by CV-DD.}
    \label{fig:imagewoof_vis}
\end{figure*}

\begin{figure*}[!t]
    \centering
    \includegraphics[width=0.9\textwidth]{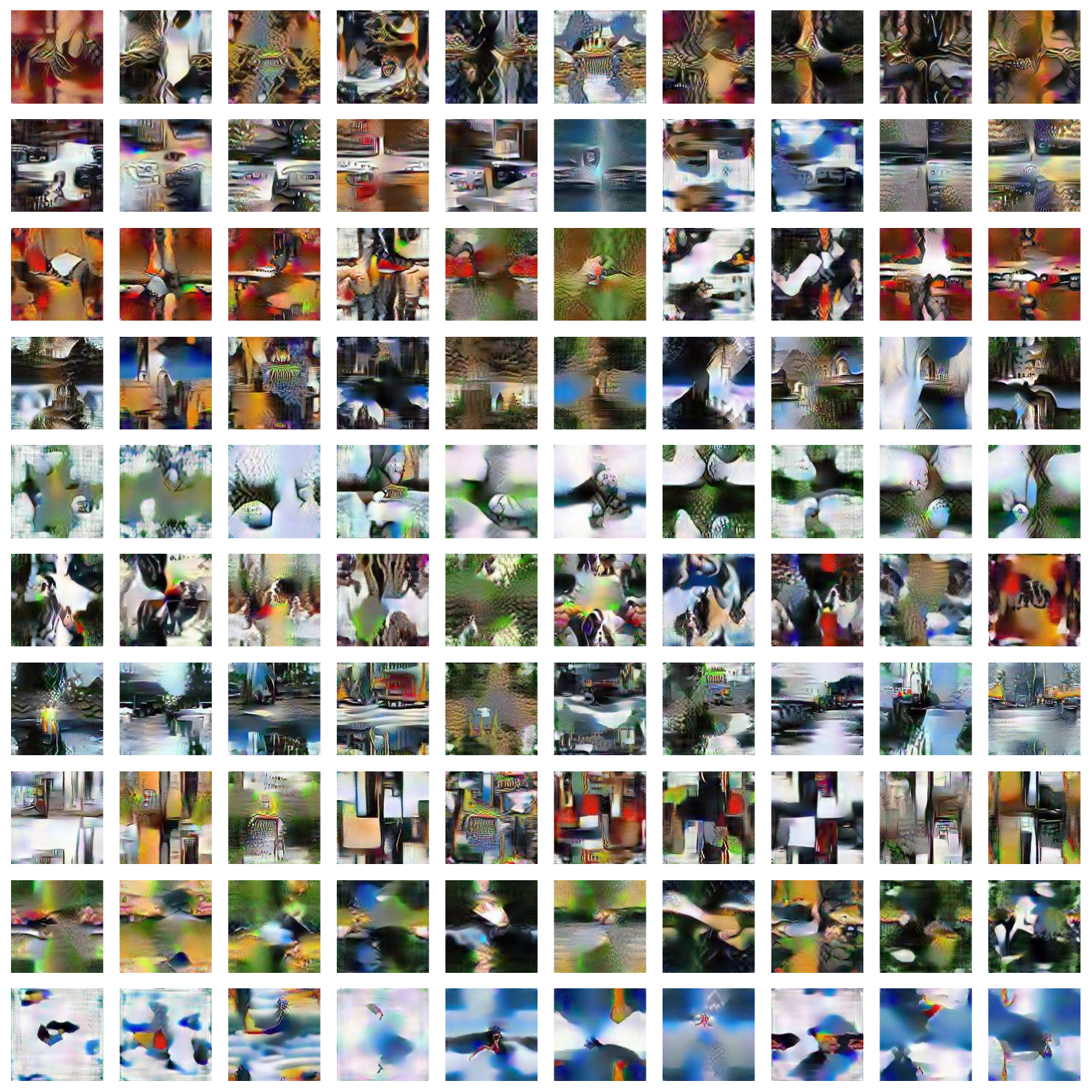}

    \caption{Visualization of synthetic data on ImageNette generated by CV-DD.}
    \label{fig:imageNette_vis}
\end{figure*}

\begin{figure*}[!t]
    \centering
    \includegraphics[width=0.9\textwidth]{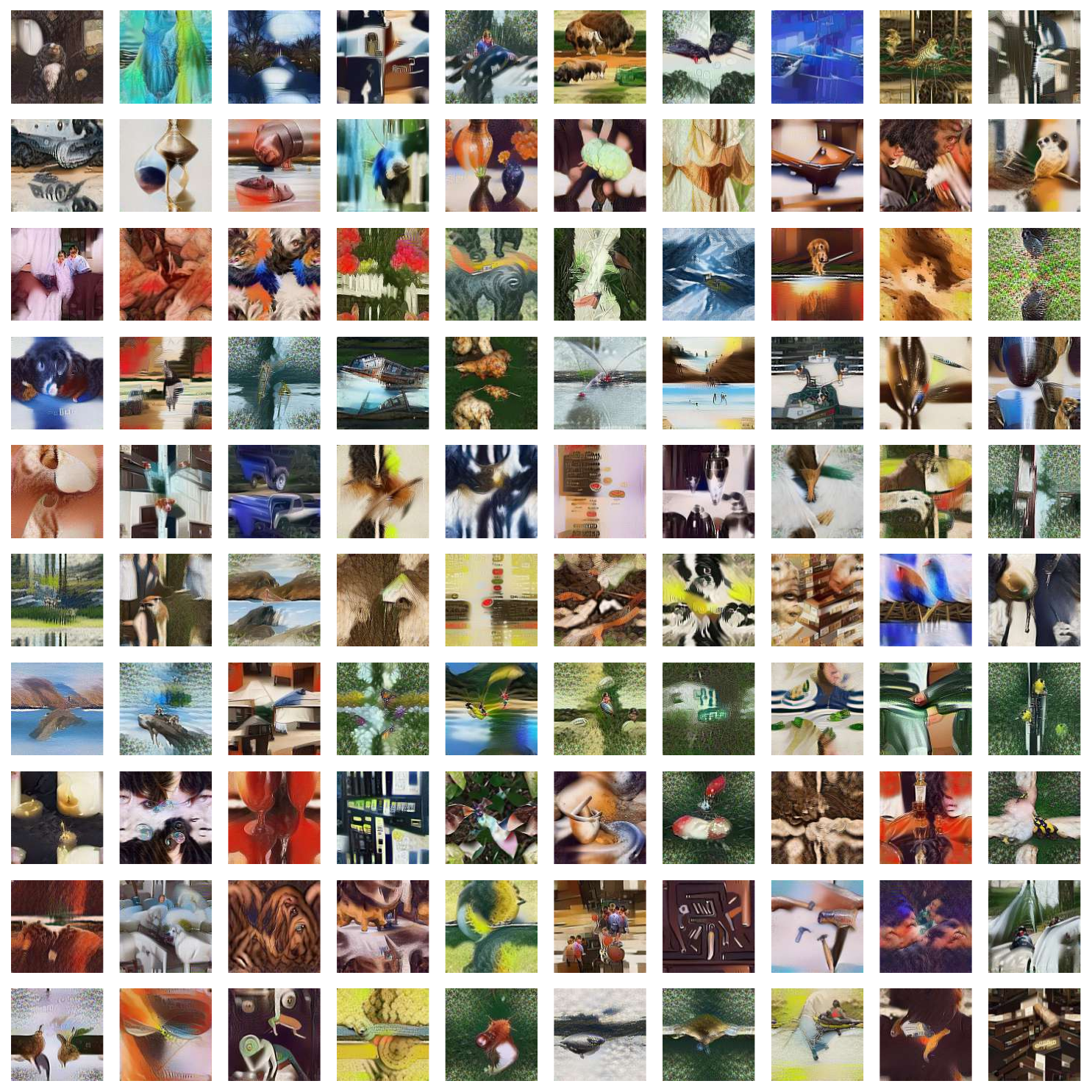}
    \caption{Visualization of synthetic data on ImageNet-1K generated by CV-DD.}
    \label{fig:imageNet1k_vis}
\end{figure*}

Extended qualitative results showcasing the visual characteristics of the distilled data synthesized by \textbf{CV-DD} are provided in Fig.~\ref{fig:cifar10_vis} (CIFAR-10), Fig.~\ref{fig:cifar100_vis} (CIFAR-100), Fig.~\ref{fig:tiny_vis} (Tiny-ImageNet), Fig.~\ref{fig:imagewoof_vis} (ImageWoof), Fig.~\ref{fig:imageNette_vis} (ImageNette), and Fig.~\ref{fig:imageNet1k_vis} (ImageNet-1K). These visualizations illustrate the diversity, class discriminability, and structural fidelity of the synthesized samples across datasets of varying resolution and semantic complexity, thereby offering intuitive evidence of the effectiveness of the proposed distillation framework.

\end{document}